\pdfoutput=1

\documentclass[11pt]{article}

\usepackage{acl}

\usepackage{natbib}
\usepackage{xcolor}
\usepackage{hyperref}
\usepackage{times}
\usepackage{latexsym}
\usepackage{multirow}
\usepackage{multicol}
\usepackage{graphicx}
\usepackage{siunitx}
\usepackage[most]{tcolorbox}

\usepackage[textsize=scriptsize]{todonotes}
\usepackage{inconsolata}

\usepackage[T1]{fontenc}

\usepackage[utf8]{inputenc}

\usepackage{microtype}

\usepackage{booktabs} 
\usepackage{amsmath}
\usepackage{mathtools}
\usepackage{multirow}
\usepackage{siunitx}
\usepackage{lineno}
\usepackage{enumitem}

\usepackage{amsmath}
\usepackage{amssymb}
\usepackage[noend]{algpseudocode}
\usepackage{algorithm}
\usepackage{array} 

\definecolor{todoblue}{RGB}{33,150,243} 

\definecolor{pink-color}{RGB}{255,1,97} 

\definecolor{red-color}{RGB}{255,111,105} 
\definecolor{dark-red-color}{RGB}{174,0,1} 
\definecolor{light-red-color}{RGB}{255,152,150}

\definecolor{green-color}{RGB}{136,216,176} 
\definecolor{dark-green-color}{RGB}{49,163,84} 
\definecolor{light-green-color}{RGB}{136,216,176} 
\definecolor{sea-green-color}{RGB}{95,158,160} 

\definecolor{dark-blue-color}{RGB}{49,130,189} 
\definecolor{light-blue-color}{RGB}{158,202,225} 

\definecolor{grey-color}{RGB}{167,173,186}
\definecolor{dark-grey-color}{RGB}{79,91,102}
\definecolor{gold-color}{RGB}{247,205,15} 

\newcommand{\redtext}[1]{\textcolor{red-color}{#1}}

\newcommand{\darkredtext}[1]{\textcolor{dark-red-color}{#1}}

\newcommand{\lightredtext}[1]{\textcolor{light-red-color}{#1}}

\newcommand{\darkgreentext}[1]{\textcolor{dark-green-color}{#1}}

\newcommand{\darkbluetext}[1]{\textcolor{dark-blue-color}{#1}}

\newcommand{\lightbluetext}[1]{\textcolor{light-blue-color}{#1}}

\newcommand{\greytext}[1]{\textcolor{grey-color}{#1}}

\newcommand{\goldtext}[1]{\textcolor{gold-color}{#1}}

\newcommand{\ours}{\mbox{\textsc{SynthesizRR}}}
\newcommand{\oursshort}{\mbox{\textsc{SynzthRR}}}

\newcommand{\gold}{\mbox{\textsc{Gold}}}
\newcommand{\fewgen}{\mbox{\textsc{FewGen}}}

\newcommand{\Products}{\mbox{\textsc{Products}}}

\newcommand{\RealNews}{\mbox{\textsc{RealNews}}}

\newcommand{\RealNewsDominant}{\mbox{\textsc{RealNews/Dominant}}}
\newcommand{\RealNewsDom}{\mbox{\textsc{RealNews/Dom}}}
\newcommand{\RNDom}{\mbox{\textsc{RN/Dom}}}

\newcommand{\RealNewsRegional}{\mbox{\textsc{RealNews/Regional}}}
\newcommand{\RealNewsReg}{\mbox{\textsc{RealNews/Reg}}}
\newcommand{\RNReg}{\mbox{\textsc{RN/Reg}}}

\newcommand{\RealNewsIndia}{\mbox{\textsc{RealNews/India}}}
\newcommand{\RealNewsInd}{\mbox{\textsc{RealNews/Ind}}}
\newcommand{\RNInd}{\mbox{\textsc{RN/Ind}}}

\newcommand{\ToI}{\mbox{\textsc{ToI}}}

\newcommand{\RetrICL}{\mbox{\textsc{RetrICL}}}
\newcommand{\NoRetrICL}{\mbox{\textsc{Non-RetrICL}}}

\newcommand{\TinyBERT}{\mbox{\textsc{TinyBERT}}}
\newcommand{\DistilBERT}{\mbox{\textsc{DistilBERT}}}
\newcommand{\BERT}{\mbox{\textsc{BERT}}}
\newcommand{\DeBERTa}{\mbox{\textsc{DeBERTa-v3L}}}
\newcommand{\DeBERTaLarge}{\mbox{\textsc{DeBERTa-v3-Large}}}

\newcommand{\GPTTwoXL}{\mbox{\textsc{GPT2-XL}}}

\newcommand{\ChatGPT}{\mbox{\textsc{GPT3.5-Turbo}}}
\newcommand{\ChatGPTShort}{\mbox{\textsc{GPT3.5-T}}}

\newcommand{\GPTDavinci}{\mbox{\textsc{GPT3.5}}}

\newcommand{\ZeroGen}{\mbox{\textsc{ZeroGen}}}
\newcommand{\ProGen}{\mbox{\textsc{ProGen}}}
\newcommand{\SunGen}{\mbox{\textsc{SunGen}}}
\newcommand{\ReGen}{\mbox{\textsc{ReGen}}}
\newcommand{\LetsSynth}{\mbox{\textsc{S3}}}
\newcommand{\AttrPrompt}{\mbox{\textsc{AttrPrompt}}}
\newcommand{\AttrPromptShort}{\mbox{\textsc{AttPmt}}}

\newcommand{\dataxhat}{\ensuremath{\mathbin{ \tilde{x} }}}
\newcommand{\dataxihat}{\ensuremath{\mathbin{ \tilde{x}^{i}} }}

\newcommand{\datayi}{\ensuremath{\mathbin{ y^{i}} }}

\newcommand{\lm}{\ensuremath{\mathcal{M}_{\text{LM}}}}

\newcommand{\studentlm}{\ensuremath{\mathcal{M}_{\text{S}}}}

\newcommand{\synthd}{\ensuremath{\mathbin{
    \mathcal{D}_{\textsc{Synth}}
}}}
\newcommand{\synthdataset}{\ensuremath{\mathbin{
    \synthd = \left\{ (\dataxihat, \datayi) \right\}^{m}_{i=1}
}}}

\newcommand{\higherbetter}{\ensuremath{(\uparrow)}}
\newcommand{\lowerbetter}{\ensuremath{(\downarrow)}}

\newcommand{\noeval}{\footnotesize{\rotatebox[origin=c]{-90}{\hspace{-0.5ex}$\otimes$}}}
\newcommand{\norelease}{\scriptsize{\rotatebox[origin=c]{-90}{\hspace{-1ex}$\bowtie$}}}

\newcommand{\insection}[2][.]{{\setlength{\parskip}{6pt} \noindent\textbf{#2#1}}}

\newcommand{\ignore}[1]{}
\newcommand{\ig}[1]{}

\newcommand{\promptsubsection}[1]{
\setlength{\parskip}{6pt} \noindent\textbf{{#1}:}
}
\newcommand{\param}[1]{
\textcolor{pink-color}{\scriptsize{\texttt{\detokenize{#1}}}}
}
\newcommand{\paramnorm}[1]{
\textcolor{pink-color}{\small{\texttt{\detokenize{#1}}}}
}
\newcommand{\prompttext}[1]{\textcolor{dark-grey-color}{#1}}

\newtcolorbox[list inside=prompt,auto counter,number within=section]{prompt}[1][]{
    colbacktitle=black!60,
    coltitle=white,
    fontupper=\footnotesize,
    boxsep=5pt,
    left=0pt,
    right=0pt,
    top=0pt,
    bottom=0pt,
    boxrule=1pt,
    #1,
}

\newcommand{\urlsmall}[1]{{\scriptsize{\url{\detokenize{#1}}}}}

\newcolumntype{h}{>{\setbox0=\hbox\bgroup}c<{\egroup}@{}}
\newcolumntype{C}[1]{>{\centering\arraybackslash}m{#1}}
\newcolumntype{R}[1]{>{\raggedleft\arraybackslash}p{#1}}
\newcolumntype{L}[1]{>{\raggedright\arraybackslash}p{#1}}

\newcommand{\cellhalign}[1]{\multicolumn{1}{c}{#1}}

\newcommand{\best}[1]{\textbf{#1}}

\newcommand{\LLaMaShortNonSc}{\mbox{LLaMa}}

\newcommand{\LLaMa}{\mbox{\textsc{LLaMa2}}}
\newcommand{\LLaMaFull}{\mbox{\textsc{LLaMa-2 Chat 13B}}}
\newcommand{\LLaMaFullNonSc}{\mbox{LLaMa-2 Chat 13B}}

\newcommand{\ClaudeShortNonSc}{\mbox{Claude}}

\newcommand{\Claude}{\mbox{\textsc{ClaudeV1}}}
\newcommand{\ClaudeFull}{\mbox{\textsc{Claude Instant-v1}}}

\newcommand{\IMDb}{\mbox{\textsc{IMDb}}}

\newcommand{\SST}{\mbox{\textsc{SST-2}}}

\newcommand{\Yelp}{\mbox{\textsc{Yelp}}}

\newcommand{\CMUMovies}{\mbox{\textsc{CMU Movie Summary}}}
\newcommand{\CMUMoviesShort}{\mbox{\textsc{Movies}}}
\newcommand{\CMUMoviesSummary}{\mbox{\textsc{Movie Summary}}}

\newcommand{\AGNews}{\mbox{\textsc{AG News}}}
\newcommand{\AG}{\mbox{\textsc{AG.}}}

\newcommand{\Hyperpartisan}{\mbox{\textsc{Hyperpartisan}}}

\newcommand{\Hyper}{\mbox{\textsc{Hyper.}}}
\newcommand{\Hyp}{\mbox{\textsc{Hyp.}}}

\newcommand{\ToIHeadlines}{\mbox{\textsc{ToI Headlines}}}

\newcommand{\Category}{\mbox{\textsc{Category}}}
\newcommand{\Categ}{\mbox{\textsc{Categ.}}}
\newcommand{\Cat}{\mbox{\textsc{Cat.}}}

\newcommand{\Humor}{\mbox{\textsc{Humor}}}
\newcommand{\Hum}{\mbox{\textsc{Hum.}}}

\newcommand{\Polarity}{\mbox{\textsc{Polarity}}}
\newcommand{\Polar}{\mbox{\textsc{Polar.}}}
\newcommand{\Pol}{\mbox{\textsc{Pol.}}}

\newcommand{\BM}{\mbox{\textsc{BM25}}}
\newcommand{\Contriever}{\mbox{\textsc{Contriever}}}
\newcommand{\ContrieverShort}{\mbox{\textsc{Contr.}}}

\newcommand{\genex}[7]{
\begin{table*}[!t]
\centering
\tiny{
\setlength{\tabcolsep}{3pt}
\begin{tabular}{C{0.12\textwidth}p{0.84\textwidth}}
\toprule
\bf{Class} & \cellhalign{\bf{Example}} \\
\midrule
\textit{(Retrieved document)} & #3
\vspace{1.5ex}
\\ 
#2 & #4 \\
\midrule
\textit{(Retrieved document)} & #6
\vspace{1.5ex}
\\ 
#5 & #7 \\
\bottomrule
\end{tabular}
\vspace{-1ex}
}
\caption{
Generated examples for #1{} task using \ours{}.
}
\vspace{-3ex}
\end{table*}
}

\title{{\ours}: Generating Diverse Datasets with Retrieval Augmentation}

\author{
Abhishek Divekar\textsuperscript{$\spadesuit$$\diamondsuit$}\thanks{Work completed while at Amazon.} \hspace{.3cm} 
Greg Durrett\textsuperscript{$\diamondsuit$} \\
\textsuperscript{$\spadesuit$}Amazon \\ 
\textsuperscript{$\diamondsuit$}Department of Computer Science, The University of Texas at Austin \\
\texttt{adivekar@amazon.com} \hspace{.3cm} \texttt{gdurrett@cs.utexas.edu} \\
}

\begin{document}
\maketitle

\begin{abstract}
It is often desirable to distill the capabilities of large language models (LLMs) into smaller student models due to compute and memory constraints. One way to do this for classification tasks is via dataset synthesis, which can be accomplished by generating examples of each label from the LLM. Prior approaches to synthesis use few-shot prompting, which relies on the LLM's parametric knowledge to generate usable examples. However, this leads to issues of repetition, bias towards popular entities, and stylistic differences from human text. In this work, we propose Synthesize by Retrieval and Refinement (\ours), which uses retrieval augmentation to introduce variety into the dataset synthesis process: as retrieved passages vary, the LLM is ``seeded'' with different content to generate its examples. We empirically study the synthesis of six datasets, covering topic classification, sentiment analysis, tone detection, and humor, requiring complex synthesis strategies. We find that \ours{}\footnote{\mbox{\footnotesize \url{https://github.com/amazon-science/synthesizrr}}} greatly improves lexical and semantic diversity, similarity to human-written text, and distillation performance, when compared to 32-shot prompting and four prior approaches.
\end{abstract}
\section{Introduction}


\begin{figure}[!t]
\centering
\includegraphics[width=0.49\textwidth]{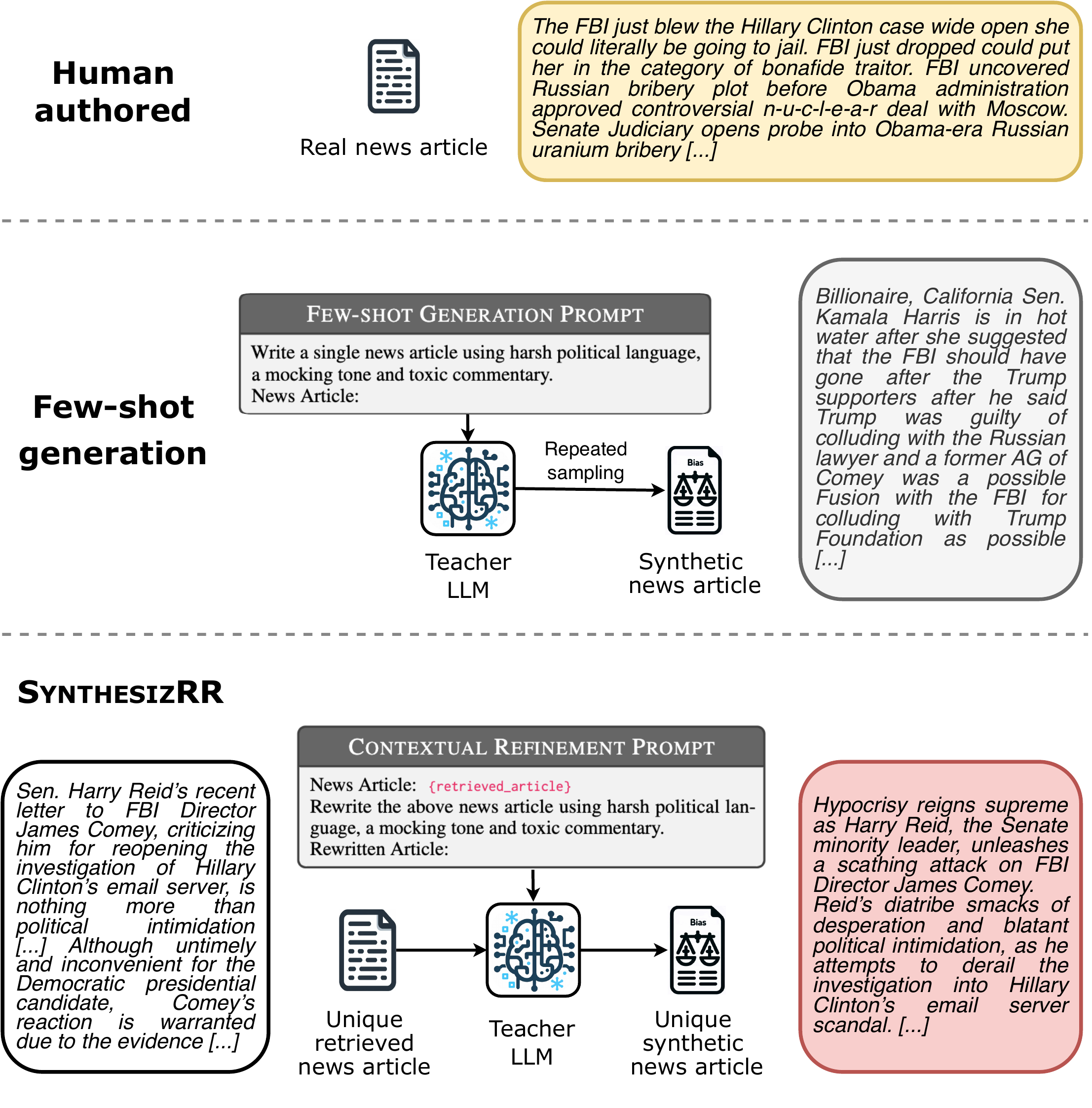}
\vspace{-2ex}
\caption{
Synthetic examples from few-shot generation (middle) and \ours{} (bottom). 
Our approach incorporates a \textit{\mbox{content sourcing}} step which retrieves documents from a corpus: for the task of detecting political bias, a news article is retrieved and the teacher LLM is prompted to produce a biased version. The resulting synthesis procedure yields diverse examples which more closely match human-written examples.
}
\vspace{-2.5ex}
\label{fig:example-political-bias}
\end{figure}
\begin{figure*}[!t]
\centering
\includegraphics[width=0.85\textwidth]{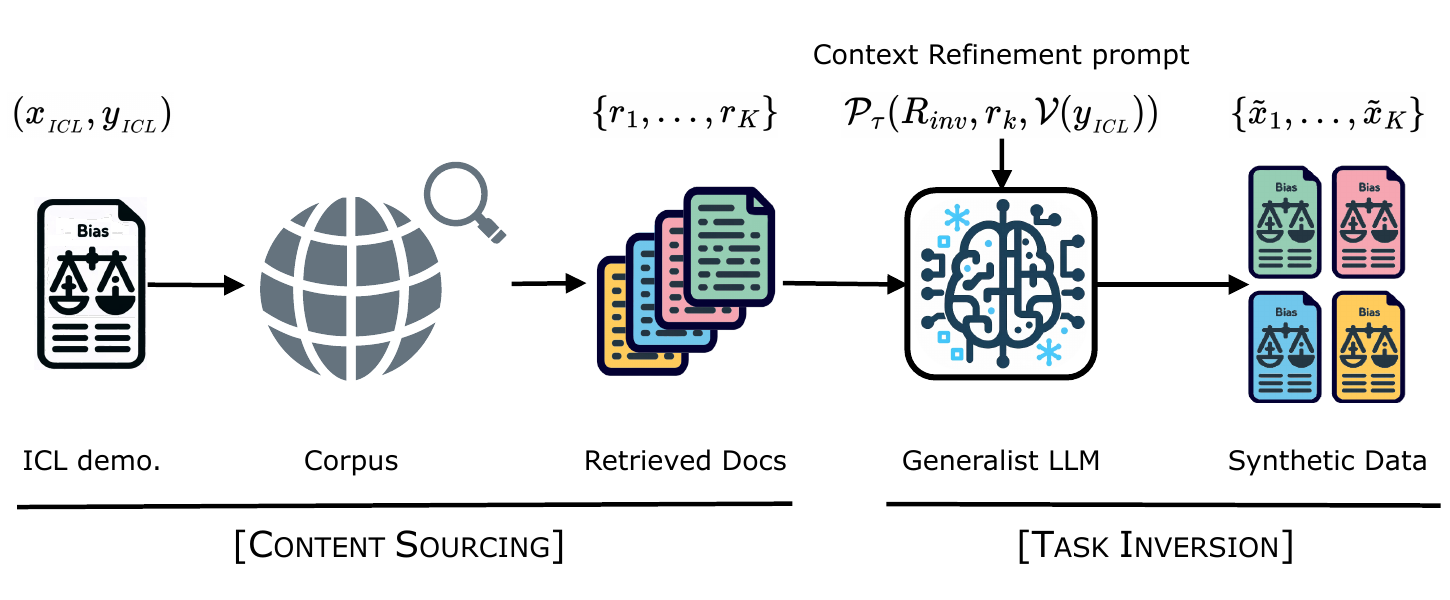}


\caption{
Abstract depiction of the \ours{} procedure. In the content sourcing stage, we retrieve $K$ unique document $\{ r_1, \dots, r_K\}$ from a large corpus for each in-context covariate $x_{\text{ICL}}$. The task-inversion stage of synthesis uses a parameterized \textit{context refinement prompt} $\mathcal{P}_{\tau}$, which takes parameters $R_{inv}$ (inversion instruction), $r_k$ (a retrieved document), and $\mathcal{V}(y_{\text{ICL}})$ (the verbalized target label). A generalist teacher LLM autoregressively generates a synthetic covariate. Each in-context example thus produces $K$ unique synthetic examples $\{\tilde{x}_1, \dots, \tilde{x}_K\}$, which we include in the dataset with target $y_{\text{ICL}}$.
}
\vspace{-2.5ex}
\label{fig:high-level}
\end{figure*}

Large Language Models (LLMs) such as \mbox{GPT-4} \citep{gpt4techreport,bubeck2023sparks}, \mbox{LLaMa} \citep{touvron2023llama2} and \mbox{Claude} \citep{Bai2022TrainingAH} are versatile \mbox{\textit{generalist}} models, capable of solving multiple tasks without parameter tuning via zero-shot or few-shot prompting. In comparison, previous approaches \mbox{fine-tuned} variants of BERT \citep{devlin-etal-2019-bert} on \mbox{task-specific} demonstrations, producing \mbox{\textit{specialist}} models. These smaller specialist models are more economical at inference time, but require at least thousands of examples to train. 

Recent work has sought to avoid this reliance on manually created examples by fine-tuning specialist models on \textit{synthetic} datasets via teacher-student distillation \citep{west-etal-2022-symbolic}. This has applications in classification \citep{yu2023large, Ye2022ZeroGenEZ, ye-etal-2022-progen}, human-preference alignment \citep{lee2023rlaif, Bai2022TrainingAH}, language understanding \citep{meng2022generating, schick-schutze-2021-generating}, and even tabular data \citep{borisov2022language}. However, synthetic data has limitations. As \citet{yu2023large} note, naive prompts generate texts with limited diversity and reflecting biases of the teacher LLMs. 

\autoref{fig:example-political-bias} illustrates the 
few-shot synthesis approach \citep{Ye2022ZeroGenEZ, ye-etal-2022-progen, yehudai2024achieving}, which we refer to as \fewgen, for the task of detecting politically-biased articles. 
With a suitable prompt and in-context examples, sampling continuations from an LLM generates plausible news in the biased style we seek to detect. However, as thousands of completions are sampled from a fixed prompt, we observe repetition, bias towards popular entities, and stylistic differences from human-written texts. Specialist models distilled from such low diversity datasets may not learn the task well. 

In this work, we seek to alleviate the lack of diversity in synthetic data. We suggest that dataset synthesis may be decomposed as two distinct LLM competencies: \textit{content sourcing}, where the LLM obtains relevant information for the task, and \textit{task inversion}, where the LLM generates a synthetic input using a target-conditioned prompt. Prior work has focused mainly on task inversion, while implicitly using the LLM's parametric memory for content sourcing. In contrast, we investigate the importance of an explicit content sourcing stage. 

We propose \textit{Synthesize by Retrieval and Refinement} (\ours), 
an example synthesis procedure guided by a retrieval corpus. 
In the content sourcing step, we use in-context learning covariates as retrieval queries to extract dozens of documents per query from a domain-specific corpus. Subsequently, a generalist LLM performs \textit{task inversion} on each retrieved document. As each prompt uses a unique retrieved document, our synthesis procedure generates diverse examples, enriched with a broad spectrum of real-world entities and assertions. 
\ig{EARLIER VERSION: In the initial step, we repurpose each in-context learning input as a retrieval query. We use this to retrieve several documents from a large domain-specific corpus. 
We then use a generalist LLM for \textit{task inversion} on each retrieved document. 
As a unique retrieved document is used in each prompt, the synthesis procedure yields more diverse examples, containing a wider range of real-world entities and assertions.  
}

We benchmark \ours{} against \fewgen{} on six text classification tasks, selected carefully to measure a variety of different styles 
of dataset synthesis. Our experiments (\S\ref{sec:expts}) reveal that \ours{} significantly surpasses \fewgen{} in diversity 
and resemblance to human-authored texts, even though both procedures utilize the same frozen LLM. In \S\ref{sec:student}, we see that student classifiers fine-tuned on \ours{}-generated data perform better than those fine-tuned on \fewgen{}. 
Finally, in \S\ref{sec:detailed-baseline-comparison}, we compare \ours{} to four state of the art approaches for synthesis of classification datasets, and find \ours{} gives higher diversity datasets, better matching human-written instances, and leads to higher student accuracy in most cases. 

Our contributions are as follows: (1) we propose a new method of example synthesis for teacher-student distillation, which grounds the task inversion step using a retrieval corpus; (2) we introduce the \ours{} \RetrICL{} algorithm to create a realistic in-context learning set for our method; (3) we empirically analyze the synthesis of six challenging classification tasks, comparing our method's textual diversity and similarity and downstream task accuracy to existing approaches; (4) we pinpoint factors affecting the quality of our synthetic datasets by varying the amount of supervised data, corpus relevance to task, number of in-context examples, and sparse vs. dense retrieval. 

\section{Background and Task setup}
\label{sec:tasks-and-complexity}

In this paper, we focus on generating datasets for challenging text classification tasks. Denote an example as consisting of input text $x$ and output $y \in \mathcal{Y}$ for output space $\mathcal{Y}$ of $C$ classes. Our goal is to produce a synthetic dataset \synthdataset{} and train a specialist language model \studentlm{} (e.g. a BERT-style pre-trained model \citep{devlin-etal-2019-bert}). We create \synthd{} via \textit{task inversion}: repeatedly prompting a teacher language model \lm{} to generate synthetic covariates $\tilde{x}$ given corresponding labels $y$. We denote the \textit{student's} task (predicting $y$ from $x$) as $\tau$ and the \textit{teacher's} task (generating $x$ given $y$) as $\tau_{inv}$.

\ours{} aims to address the lack of diversity by leveraging retrieval during the content sourcing step. 
We assume the existence of a corpus $\mathcal{R}$ where each document may hold task-relevant information. However, documents need not originate from the same distribution as our task covariates; even distantly related documents can yield valuable synthetic examples. For instance, we shows that we can successfully generate reviews and humorous questions from a corpus of product descriptions. We also assume access to a \emph{seed set} of examples $\mathcal{D}_{\textsc{Seed}} = \{(x_1,y_1),\ldots,(x_n,y_n)\}$ which is sufficiently large to represent the classes but small enough to be manually compiled by a user in a few hours; in experiments, we use the in-context learning set as $\mathcal{D}_{\textsc{Seed}}$. Importantly, we assume the seed set is insufficient to train an effective student, and a larger \synthd{} ($m >> n$) is needed. 

\autoref{fig:high-level} illustrates our method for generating distributionally similar covariates. Initially, we retrieve documents based on the examples in $\mathcal{D}_{\textsc{Seed}}$, assuming that the corpus contains sufficient domain-similar documents. We then construct a \textit{context refinement} instruction to perform task inversion on each retrieved document. This approach provides the LLM with a unique and grounded prompt for each generated example, thereby circumventing the need for the teacher LLM to memorize extensive corpus data within its limited parameters. Task inversion may be challenging due to the mismatch between retrieved documents and test examples; to overcome this, we limit our investigation to teacher LLMs demonstrating strong instruction-following capabilities \citep{ouyang2022rlhf, touvron2023llama2, Bai2022TrainingAH}.




\begin{algorithm}[t!]
\small
\caption{SynthesizRR \RetrICL}
\begin{algorithmic}

\State {\textbf{Input} A set of seed examples $\mathcal{D}_{\textsc{Seed}}$, retrieval corpus $\mathcal{R} = \{r_k\}$, retrieval model $\mathcal{M}_{\mathrm{ret}}$, expansion factor $K$, cosine-similarity criterion $(s_{\alpha}, s_\beta)$, teacher model \lm, prompt template $\mathcal{P}_{\tau}$, context refinement instruction $R_{inv}$, verbalizer \mbox{$\mathcal{V}: \{y_1, \dots, y_{C}\} \rightarrow \{v_1, \dots, v_{C}\}$}}.

\State {\textbf{Output} Synthetic dataset $\synthd{}$}

\State {\textbf{Procedure} $\textsc{SynthesizRR}(\mathcal{D}_{\textsc{Seed}}, \mathcal{R})$:}

\State $\mathcal{D}_{\textsc{Retr}} \leftarrow \emptyset$
\State $\mathcal{D}_{\textsc{ICL}} \leftarrow \emptyset$
\State $\synthd{} \leftarrow \emptyset$
\State $\triangleright$ {Content sourcing using retrieval:}
\For{$(x,y) \in \mathcal{D}_{\textsc{Seed}}$}
\State $[r_1,\ldots,r_K] \leftarrow \mathcal{M}_{\mathrm{ret}}(x)$
\State $\Gamma_{K} \leftarrow [r_1,\ldots,r_K]$
\State $\mathcal{D}_{\textsc{Retr}} \leftarrow \mathcal{D}_{\textsc{Retr}} \cup \{(x, y, \Gamma_{K})\}$
\EndFor
\State $\triangleright$ {In-context learning set construction:}
\For{$(x, y, \Gamma_{K}) \in \mathcal{D}_{\textsc{Retr}}$}
\For{$r_k \in \Gamma_{K}$}
    \State $ \mathcal{D}_{\textsc{ICL}} \leftarrow \mathcal{D}_{\textsc{ICL}} \cup \{(r_k, x)\}$ \textbf{if} 
$s_{\alpha} \le \cos(x, r_k) \le s_\beta$
    \EndFor
\EndFor
\State $\triangleright$ {Task inversion:}
\For{$(x, y, \Gamma_{K}) \in \mathcal{D}_{\textsc{Retr}}$}
\For{$r_k \in \Gamma_{K}$}
    \State $\mathcal{D}_{\textsc{shots}} \underset{\raisebox{2pt}{$\scriptscriptstyle $}}{\sim} \mathcal{D}_{\textsc{ICL}}$
    \For{$j \in [1, \dots] \text{ \textbf{ until } } \dataxhat^{i}_{j} = \text{<eos>}$}
        \State $\dataxhat^{i}_{j} { }{ \sim }{ } \lm\left( \cdot | \dataxhat^{i}_{<j}, \mathcal{P}_{\tau}(R_{inv}, r_k, \mathcal{V}(y)), \mathcal{D}_{\textsc{shots}} \right)$
    \EndFor
  \State $\synthd{} \leftarrow \synthd{} \cup \{ (\tilde{x}^{i}, y)\}$
\EndFor
\EndFor
\State \Return $\synthd{}$
\end{algorithmic}
\label{alg:synthesizrr}
\end{algorithm}

\section{Method}
\label{sec:method}



Algorithm~\ref{alg:synthesizrr} shows our dataset generation method. We distill a student model in these steps:

\insection[:]{Step 1. Content sourcing using retrieval} \ours{} uses each in-context covariate $x_{{}_{\text{ICL}}}$ as a query for information retrieval, in addition to its subsequently role during in-context learning. For each query, we retrieve $K$ documents $\Gamma_{K} = [r_1, \dots, r_K]$ of progressively decreasing cosing similarity using the dense retriever $\mathcal{M}\mathrm{ret}$. We retain documents with cosine similarity in \mbox{(0.4, 0.9),} to ensure minimum similarity while excluding overly similar documents as potential duplicates of $x_{{}_{\text{ICL}}}$. Each resulting triplet $(x_{{}_{\text{ICL}}}, y_{{}_{\text{ICL}}}, \Gamma_{K})$ is appended to set $\mathcal{D}_{\textsc{Retr}}$.




\insection[:]{Step 2. In-context set construction} The subsequent task inversion step also benefits from in-context demonstrations, but it is challenging to construct demonstrations which effectively captures our context refinement task $r_{k}^{i} \rightarrow \dataxhat^{i}$. 
We explored two approaches to in-context learning. 

\textbf{1. \RetrICL:} we use retrieval to construct a set of ICL examples $\mathcal{D}_{\textsc{ICL}}$, such that each ICL example mirrors the format of our task-inversion prompts. We select top-1 and top-2 retrieved results from the densely retrieved results, and use a cosine-similarity criterion $s_{\alpha} \le \cos(x_{{}_{\text{ICL}}}, r_k) \le s_\beta$ to asses the potential match between the retrieved document $r_k$ and $x_{{}_{\text{ICL}}}$. 
Although the in-context pair may not match exactly, they demonstrate the required format as per \autoref{sec:prompts}. 

\textbf{2. \NoRetrICL:} a baseline method, which uses retrieval for content sourcing, but not for in-context learning. For each generation we select $N=32$ ICL examples at random from $\mathcal{D}_{\textsc{Seed}}$. Each example is appended with a prefix like \textit{``News Article:''} or \textit{``Product details:''} but we do \textit{not} add the context refinement instruction. After the ICL examples, we append the retrieved document $r_k$ and context refinement instruction $R_{inv}$ to form the final prompt. This format closely mirrors the in-context learning prompt used by \fewgen{}, but also incorporates content-sourcing elements $r_k$ and $R_{inv}$. This baseline highlights the value added by constructing $\mathcal{D}_{\textsc{ICL}}$ in the \RetrICL{} approach.


\insection[:]{Step 3. Task inversion using context refinement} 
The minimum elements of a task inversion prompt $\mathcal{P_{\tau}}$ are the context refinement instruction $\mathcal{I}_{inv}$ and \mbox{target} $y$. 
We use a verbalizer function $\mathcal{V}$ \citep{schick-schutze-2021-generating, van-de-kar-etal-2022-dont} to provide a unique text representation of each label, i.e. \mbox{$\mathcal{V}: \mathcal{Y} \rightarrow \{v_1, \dots, v_{C}\}$}. We follow prior work on classification-based task inversion \citep{schick-schutze-2021-generating, Ye2022ZeroGenEZ, ye-etal-2022-progen, yu-etal-2023-regen, gao2023selfguided} and use descriptive verbalizations to induce label-separability in the final dataset.

\fewgen{} uses the standard causal language modeling objective to induce next-token probabilities from teacher LLM, \lm. Nucleus sampling \cite{Holtzman2020CuriousCase} is used to autoregressively sample next tokens until the \url{<eos>} token is generated. This becomes synthetic example $\dataxhat^{i}$.

\vspace{-3ex}
\begin{flalign}
\mbox{$\dataxhat^{i}_{j} \underset{\raisebox{2pt}{$\scriptscriptstyle p$}}{\sim} \lm\left(\cdot | \dataxhat^{i}_{<j}, \mathcal{P}_{\tau}(I_{inv}, \mathcal{V}(y)) \right)$}
\end{flalign}
\vspace{-3ex}

For each label $y$, we fix this prompt and sample $m/C$ times to generate the synthetic dataset. 

In \ours{}, we create the synthetic dataset from each triplet in $\mathcal{D}_{\textsc{Retr}}$.
The retrieved documents $\Gamma_{K} = [r_1, \dots, r_K]$ have lexical and semantic overlap with the query $x_{{}_{\text{ICL}}}$. However, corpus documents may be distributionally dissimilar from real task covariates, due to the nature of documents or chunking process \citep{mialon2023augmented}. To address this, we use \lm{} to perform task inversion from the content of each retrieved document, a process we refer to as \textit{contextual refinement}. $\mathcal{P_{\tau}}$ is thus composed from the contextual refinement instruction $\mathcal{R}_{inv}$, each document $r_k \in \Gamma_{K}$, and the verbalized target for the query, i.e. $\mathcal{V}(y_{{}_{ICL}})$. The LLM's context window thus sees a unique and grounded prompt when auto-regressively generating each synthetic input $\dataxhat^{i}$:

\vspace{-3ex}
\begin{flalign}
\mbox{$\dataxhat^{i}_{j} \underset{\raisebox{2pt}{$\scriptscriptstyle p$}}{\sim} \lm\left( \cdot | \dataxhat^{i}_{<j}, \mathcal{P}_{\tau}(R_{inv}, r_k, \mathcal{V}(y_{{}_{ICL}})) \right)$}, 
\end{flalign}
for all documents $r_k \in \Gamma_{K}$. We continue to use nucleus sampling to get diverse generations. Each original in-context example thus produces $K$ unique synthetic examples $\{\tilde{x}_1, \dots, \tilde{x}_K\}$; we call $K$ the ``expansion factor''. To promote adherence to $\mathcal{R}_{inv}$, we sample pairs from $\mathcal{D}_{\textsc{ICL}}$ to create in-context examples following the same format. Our final dataset is constructed as:\\ $\mathcal{D}_{\textsc{Synth}} = \bigcup\limits_{(x, y, \Gamma_{K}) \in \mathcal{D}_{\textsc{Retr}}} { } \bigcup\limits_{r_k \in \Gamma_{K}} \left\{ (\dataxihat, y) \right\}$.

\insection[:]{Step 4. Student distillation} 
The student is fine-tuned on \synthd{} by passing the BERT \url{[CLS]} token embedding of $\dataxhat$ through a feedforward layer. This produces a probability distribution over the label space $C$. We optimize the cross-entropy loss of the true label $y$. As we derive $\dataxhat$ from a teacher LLM, this can be considered a form of symbolic knowledge distillation \citep{west-etal-2022-symbolic}. 
\begin{table}[!t]
\centering
\footnotesize{
\setlength{\tabcolsep}{2pt}
\begin{tabular}{
C{60pt}
h
R{20pt}
r
C{44pt}
h
C{32pt}
}
\toprule
\textbf{Dataset}        
& \textbf{Task}    
& \textbf{Class}
& \textbf{Train, Test}
& \textbf{Corpus} 
& \textbf{Content sourcing}
& \textbf{Difficulty}
\\ 
\midrule
\small{\AGNews}                 
& News Topic   
& 4
& $115\text{k}, 7.6\text{k}$   
& \small{\RNDom}
& Easy
& Easy
\\
\small{\ToIHeadlines}
& News Topic   
& 10
& $52\text{k}, 10\text{k}$
& \small{\RNInd}
& Easy
& Easy
\\
\small{\Hyperpartisan}
& Political bias   
& 2
& $516, 65$                 
& \small{\RNDom}
& Easy
& Medium
\\
\small{\Polarity}
& Review sent.
& 2
& $72\text{k}^*, 7.2\text{k}^*$
& \small{\Products}
& Hard
& Medium
\\
\small{\Category}
& Review category
& 23
& $30\text{k}^*, 2.4\text{k}^*$     
& \small{\Products}
& Hard
& Medium
\\
\small{\Humor}
& Question humor
& 2
& $15\text{k}, 3\text{k}$      
& \small{\Products}
& Medium
& Hard
\\
\small{\IMDb}
& Movie sent.
& 2
& $20\text{k}, 25\text{k}$
& \small{\CMUMoviesShort}
& Hard
& Medium
\\
\small{\SST}
& Movie sent.
& 2
& $54\text{k}, 872$
& \small{\CMUMoviesShort}
& Hard
& Medium
\\


\bottomrule
\end{tabular}

\vspace{-1ex}
}
\caption{
Dataset statistics and our estimate of task inversion difficulty. ${}^*$Downsampled for convenience.
}
\vspace{-2ex}
\label{tab:tasks}
\end{table}

\section{Experimental Setup}
\label{sec:expt-setup}

\insection{Tasks and their difficulty} We perform our main experiments on the first 6 datasets in \autoref{tab:tasks}, selected carefully to measure how the teacher LLM performs on task inversion tasks of varying difficulty. Previous work only benchmarked sentiment and topic classification datasets like \IMDb{} \cite{maas-etal-2011-learning} and \AGNews{} \citep{zhang2015character}. 
We broaden from topic classification, which primarily involves summarization during the task inversion step, which LLMs are adept at \citep{goyal2022zeroshotnews}. 
\Hyperpartisan{} \citep{kiesel-etal-2019-semeval} detects bias in political news, so the task inversion step includes a more substantial rewriting of neutral retrieved articles to form biased examples. 
\Category{} and \Polarity{} are prevalent product review tasks \citep{yu2023large, yu-etal-2023-regen, gao2023selfguided}; we generate reviews from retrieved products which must conform to categorical and sentiment classes.
Task inversion for \Humor{} \citep{humor} involves generating humorous questions from retrieved product details, which requires additional skills from the teacher. 
Prompts for all tasks are in \autoref{sec:prompts}.

\begin{table}[!t]
\centering
\footnotesize{
\setlength{\tabcolsep}{1.8pt}
\begin{tabular}{
c
c
c
C{23pt}
r
}
\toprule
\textbf{Corpus}         & \textbf{Domain}   & \textbf{Size}    & \textbf{Doc.} & \textbf{Tokens} \\ 
\midrule
\RealNewsDom    & US/EU News    & $30.1\text{M}$   & Article     & $27.1\text{B}$ \\
\RealNewsReg    & Regional News     & $2.7\text{M}$    & Article     & $2.1\text{B}$ \\
\textsc{\RealNewsInd}  & Indian News     & $0.9\text{M}$    & Article     & $0.6\text{B}$ \\
\Products   & E-commerce        & $15.0\text{M}$   & Product   & $2.3\text{B}$ \\
\CMUMoviesSummary & Movies  & $42\text{K}$  & Plot & $0.02\text{B}$ \\
\bottomrule
\end{tabular}

\vspace{-1ex}
}
\caption{
Corpus statistics with \LLaMa{} tokenizer.
}
\vspace{-3ex}
\label{tab:corpus}
\end{table}


\autoref{tab:corpus} describes corpora used for retrieval. We consider five corpora in different domains, each with varying numbers of records. Three are subsets of \RealNews{} \citep{zellers2019grover}, as described in \autoref{sec:preprocessing}: \RealNewsDominant{} (US/EU News), \RealNewsRegional{} (Regional News), \RealNewsIndia{} (Indian News). We also use \Products{} (Amazon products metadata, \citep{ni-etal-2019-justifying}) and \CMUMoviesSummary{} (movie summaries, \citep{bamman-etal-2013-learning}. 
Each task in \autoref{tab:tasks} is associated with the corpus we consider most relevant. 
In \S\ref{sec:detailed-baseline-comparison}, we compare to four prior approaches on three other tasks: \IMDb{} \cite{maas-etal-2011-learning}, \SST{} \cite{socher-etal-2013-recursive} and \AGNews. These sentiment and topic tasks are less aligned with our goals and thus excluded from our main evaluation.

\begin{table*}[!t]
\centering
\small
\setlength{\tabcolsep}{3pt}
\begin{tabular}{C{0.18\textwidth}p{0.78\textwidth}}
\toprule
\bf{Method} & \cellhalign{\bf{Example}} \\
\midrule

\gold{} & There is {decent bass, but the highs are a bit soft}. A quick tweak to my equalizer, and they're great. After reading several of the reviews on Amazon, I was a bit worried about the sound, but now that I have them I'm very happy. They're a {good} price, and sooooo much better than the {little ipod-like earbuds} I've tried before. Those never stayed in my ear, and the {bass} never made me happy. \\
\midrule
\fewgen{} & I've been a very happy customer of \greytext{this company} for a long time. It is \greytext{fast} and \greytext{does everything I need it to}. I would definitely recommend it to anyone looking for a \greytext{good} external drive. However, I do have one issue with the product. The \greytext{instructions that come with it are not very clear} and I had a hard time figuring out how to properly use it. \\
\midrule
\textit{(Retrieved Product)} & \textit{Portable Laptop Microphone. Connects to 1/8" mini microphone input on laptop. Right-angle shaped. Flat-frequency response.} \\ [0.5ex]
\ours{} & The portable laptop microphone is {right-angled} and has a {flat-frequency response}, making it easy to use for {online meetings and interviews}. It connects to the 1/8" mini microphone input on my laptop and has worked great for the past two months, but I have noticed some distortion in the audio when I move around too much. Overall, it's a great value for the price and has made my remote work and video conferencing much more productive and efficient. \\

\bottomrule
\end{tabular}
\vspace{-1ex}
\caption{
Real and synthetic examples from ``electronics'' class of \Category. \greytext{Grey text} indicates lack of specifics.
}
\vspace{-3ex}
\label{tab:real-synthetic-examples}
\end{table*}






\ig{\begin{itemize}[leftmargin=12pt, itemsep=0pt, topsep=6pt]
\item \RealNewsRegional{} is a subset of \RealNews{} \citep{zellers2019grover}. It includes 2.7M articles from non-US and non-EU websites. We manually check \RealNews{} websites and identified 141 regional-news websites with headquarters in 56 non-US and non-EU countries: India, Pakistan, Nigeria, Philippines, etc. (Appendix). 
\item \RealNewsIndia{} is further filtered to only include Indian news websites.
\item \RealNewsDominant{} is the remaining 30.1M articles from 1063 news websites headquartered in 20 countries (of which $\gt 75\%$ are US-based).

\end{itemize}
\vspace{-1ex}
}


\insection{Models} We use \Contriever{} \citep{izacard2022unsupervised} for dense retrieval from each corpus. This performs a semantic match between the query and each document using cosine-similarity. In  \autoref{sec:bm25}, we also perform an ablation study using \BM{} as a sparse retriever, which does lexical matching between each query-document pair.

As \textbf{teacher models}, we primarily use a frozen \mbox{Llama-2} Chat 13B \citep{touvron2023llama2} for the task inversion step in \ours{} and \fewgen{}. We also experiment with \ClaudeFull{} as described in \autoref{sec:hyperparams}. For in-context learning (ICL) \cite{gpt3}, we select examples randomly from the train set: 50 ICL examples/class for multi-class and 100/class for binary tasks. We believe this is a realistic number of examples that a system designer could source if they were to put some effort into building a specialist model. We explore approaches to bootstrap this seed set in limited-supervision settings \autoref{sec:bootstrap-seed}. 

Specialization performance is measured on \textbf{student LMs} \DeBERTaLarge{} (435M params, \citet{debertav3}) and \DistilBERT{} (66M params, \citet{Sanh2019DistilBERT}).


\begin{figure}[!t]
\includegraphics[width=0.47\textwidth]{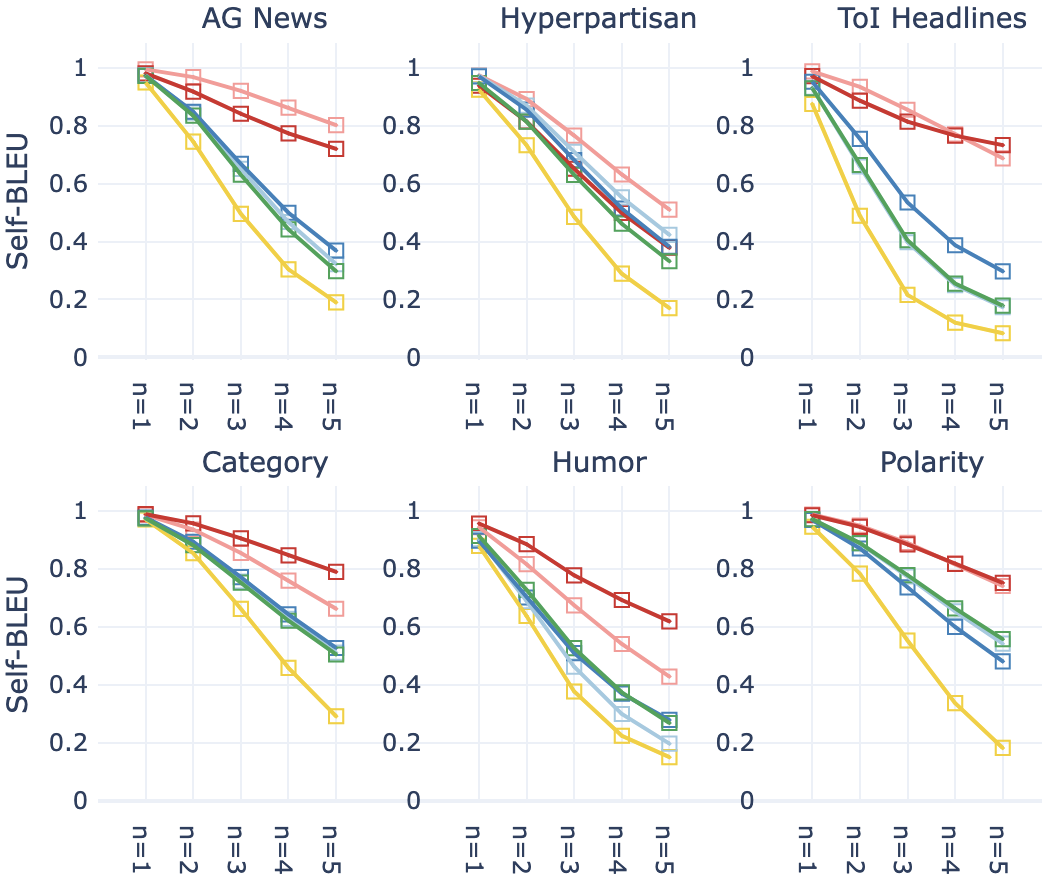}


\vspace{-0.5ex}
\centering
\caption{
Self-BLEU \lowerbetter{} for ngrams n=1-5. Comparison: \goldtext{\gold{}}, \lightredtext{\fewgen{} 0-shot}, \darkredtext{\fewgen{} 32-shot}, \lightbluetext{\ours{} 0-shot}, \darkbluetext{\ours{} 3-shot \RetrICL}, \darkgreentext{\ours{} 32-shot \NoRetrICL}.
}
\vspace{-3ex}
\label{fig:self-bleu}
\end{figure}

\insection{Evaluation criteria} Text generation can be challenging to evaluate objectively in multi-task scenarios \citep{llm-eval-survey}. Therefore in \S\ref{sec:expts} we evaluate synthetic text based on several criterion, to detect behaviours we observe during synthesis as in \autoref{tab:real-synthetic-examples}. \textbf{Self-BLEU} \citep{bleu,zhu2018texygen} measures lexical diversity of the dataset based on $n$-gram overlap between pairs of examples. \textbf{Entity entropy} measures the \textit{diversity of entities} using the probability distribution of each of 16 entity-types, inferred using spaCy's \url{en_core_web_lg} \citep{honnibal2020spacy}. Datasets which over-represent popular entities score lower on entropy. On the other hand, \textbf{Entity recall} and \textbf{Entity KL divergence} compares the \textit{similarity of entities} compared to \gold{}, and datasets which reproduce entities frequently seen in \gold{} data score higher. \textbf{MAUVE} \citep{liu-etal:divergence:neurips2021} measures similarity to human-written text by using pretrained representations from a gpt2-xl model, indicating distributional differences in the generated text. 

\begin{figure}[!t]
\includegraphics[width=0.47\textwidth]{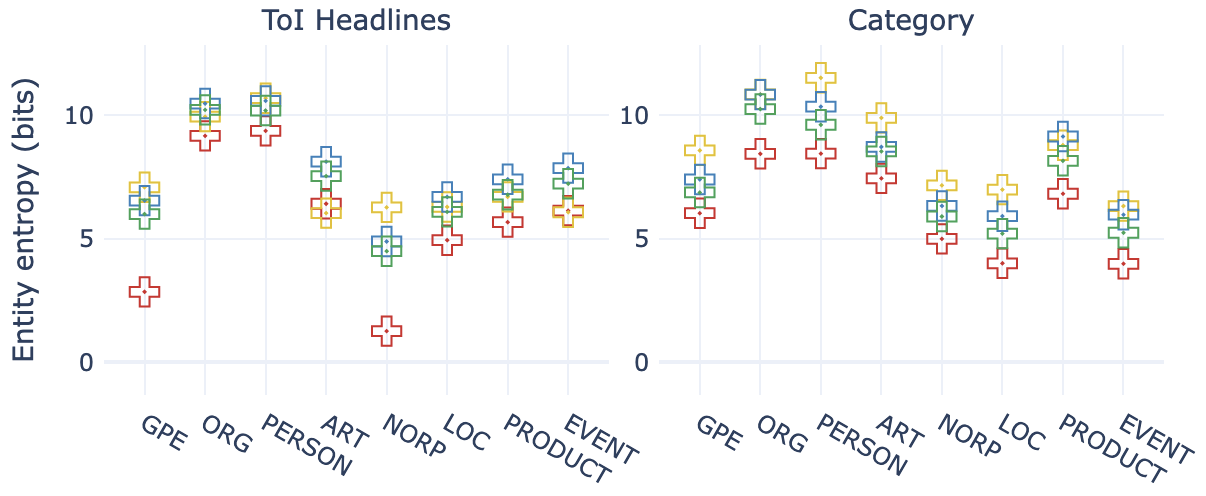}


\vspace{-0.5ex}
\centering
\caption{
Entity entropy \higherbetter{} on \ToI{} (headlines) and \Category{} (reviews). Comparison: \goldtext{\gold{}}, \darkredtext{\fewgen{} 32-shot}, \darkbluetext{\ours{} 3-shot \RetrICL} and \darkgreentext{\ours{} 32-shot \NoRetrICL}. Zero-shot results are similar for \ours{} and worse for \fewgen; we omit them. 
}
\vspace{-3ex}
\label{fig:entity-entropy}
\end{figure}
\begin{table}[!t]
\centering
\footnotesize{
\setlength{\tabcolsep}{1.9pt}

\begin{tabular}{
L{44pt}
r
r
r
r
C{30pt}
C{30pt}
}
\toprule
\bf{Method} & \textsc{Norp} &   \textsc{Org} &  \textsc{Person} &   \textsc{Gpe} &  Recall \higherbetter &  KL div. \lowerbetter \\
\midrule
\multicolumn{7}{c}{\underline{\textsc{Unique Entities}}} \vspace{1ex} \\

\gold &  319 &  3943 &   3952 &   712 &  - &  - \\
\fewgen*  & 43 &  480 &     400 &   73 &     0.05 &      - \\
\oursshort${}^{\dagger}$ & 137 &  2718 &    1528 &  238 &     \textbf{0.12} &     - \\
\oursshort${}^{\ddagger}$  & 109 &  1755 &    1012 &  178 &     0.10 &     - \\

\midrule

\multicolumn{7}{c}{\underline{\textsc{Total Entities}}} \vspace{1ex} \\

\gold & 843 &  7233 &   6096 &  1558 &  - &  - \\
\fewgen*  & 94 &  775 &     506 &   96 &     0.23 &      3.10 \\
\oursshort${}^{\dagger}$ & 319 &  3991 &    1989 &  397 &     \textbf{0.35} &     \textbf{2.35} \\
\oursshort${}^{\ddagger}$  & 314 &  2699 &    1464 &  363 &     0.32 &     2.52 \\

\bottomrule
\end{tabular}
\vspace{-1ex}
}
\caption{
Entity similarity in \Category{} ($8$K). We show the counts of unique and total entities for 4 entity-types. \textit{Entity recall} measures the fraction of \gold{} entities co-occuring in the synthetic data; in the bottom half, we additionally weigh each entity by its frequency in \gold{}. \ig{\textit{Entity KL divergence} is the KL divergence of each entity-type. }Notation: *32-shot; ${}^{\dagger}$3-shot \RetrICL{}; ${}^{\ddagger}$32-shot \NoRetrICL.
}
\vspace{-3ex}
\label{tab:entity-count}
\end{table}


\section{Results: Intrinsic Evaluation}
\label{sec:expts}

In this section, we focus on evaluating intrinsic properties of the generated datasets, including their diversity and entity coverage. 
We focus on a \LLaMaFull{} teacher LLM, retrieving from Contriever using corpora per \autoref{tab:tasks} (we analyze changing the  retrieval corpus in ~\autoref{sec:domain-shift}). 
We generate datasets of size in relation to the number of \gold{} rows: $8$K rows (\AGNews, \ToIHeadlines, \Category),  $4$K rows (\Polarity) or $2$K rows (\Hyperpartisan{}, \Humor). Example generations are in \autoref{sec:examples}.  

\insection[]{RQ: Does retrieval augmentation improve lexical diversity?}
\autoref{fig:self-bleu} shows lexical diversity within the dataset. Human-written texts (\gold) score high on lexical diversity (low Self-BLEU). \fewgen{} texts tend to reuse the same words and phrases, leading to repeated text across generations (high Self-BLEU). \ours{} text has lexical diversity approaching human text for all n-gram values. We note in-context learning has an inconsistent effect; it improves the lexical diversity for news corpora but not for products.

\insection[]{RQ: Does \ours{} address entity diversity?}
\textit{Popularity bias} is a phenomenon wherein LLM generations tend to over-represent popular ``head'' entities. This has been studied for QA tasks \citep{mallen-etal-2023-trust, Kandpal2023Large}. 


In \autoref{fig:entity-entropy} we see how \ours{} eliminates popularity bias across entity types. By sourcing from the long-tail of retrieval results ($k=50$), the generated dataset has much higher entity entropy compared to \fewgen. This positions \ours{} closer to \gold, which also shows high entity entropy.

\begin{table}[!t]
\sisetup{round-mode=places, round-precision=1} 
\centering
\setlength{\tabcolsep}{1pt}
\footnotesize{
\begin{tabular}{
l
S[table-format=3.2]
S[table-format=3.2]
S[table-format=3.2]
S[table-format=3.2]
S[table-format=3.2]
S[table-format=3.2]
}
\toprule
\textbf{Method}
& \AG           & \Hyp        & \ToI         &  \Cat     & \Hum         & \Pol \\ 
\textit{(Dataset size)} & \text{($8$K)}    & \text{($2$K)} & \text{($8$K)}  &  \text{($8$K)}   & \text{($2$K)}  & \text{($4$K)} \\ 

\midrule
\multicolumn{7}{c}{\underline{\textsc{Zero shot}}} \vspace{1ex} \\
\fewgen
&   56.57 &   53.68 &   62.79 &    \bf{63.2} &   75.6 &    62.82 \\
\oursshort
&  \bf{90.3} &  \bf{59.2} & \bf{63.0} &        61.06 &         \bf{82.9} &        \bf{78.6} \\

\midrule
\multicolumn{7}{c}{\underline{\textsc{Few shot}}} \vspace{1ex} \\
\fewgen*
&    56.7       & 65.39     & 60.33     & 65.8      & 78.14     & 69.21 \\
\oursshort${}^{\dagger}$ 
&    \bf{92.0}  & \bf{72.8} & \bf{87.9} & \bf{75.2} & \bf{87.5} & \bf{89.9} \\
\oursshort${}^{\ddagger}$ 
&    91.76      & 67.86     & 67.16     & 75.14     & 86.95     & 83.2 \\
\bottomrule

\end{tabular}
}
\vspace{-1ex}
\caption{
MAUVE similarity score ($\uparrow$) using GPT2-XL embeddings. 
Notation: *32-shot; ${}^{\dagger}$3-shot \RetrICL{}; ${}^{\ddagger}$32-shot \NoRetrICL.
}
\vspace{-3ex}
\label{tab:mauve}
\end{table}

\begin{table*}[!h]
\centering
\sisetup{round-mode=places, round-precision=1} 
\setlength{\tabcolsep}{2pt}
\footnotesize{
\begin{tabular}{
l
c
h 
S[table-format=3.2]  
S[table-format=3.2] S[table-format=3.2] 
S[table-format=3.2] S[table-format=3.2] S[table-format=3.2] 
l  
}
\toprule
\textbf{Method}
& \multirow{2}{*}{\textbf{Teacher LM}}
&  
& \AG{} 
& \Hyper{} 
& \ToI{}
& \Categ{} 
& \Humor{} 
& \Polar{} 
& \multirow{2}{*}{\textbf{Avg}}
\\
\textit{(Dataset size)}
& 
& 
& \text{($8$K)}
& \text{($2$K)}
& \text{($8$K)}
& \text{($8$K)}
& \text{($2$K)}
& \text{($4$K)}
&
\\ 
\midrule
\gold & -
& \DeBERTa{}                     &  91.015789 &  93.230769 &  82.488 &  81.516667 &  93.06 &  95.2645  & 89.43
\\
\textsc{Seed} & -
& \DeBERTa{}                     &  83.9 &  82.5 &  67.5 &  71.7 & 85.0  & 90.9   & 80.25
\\


\toprule
\multicolumn{10}{c}{\underline{\textsc{Zero-shot}}} \vspace{1ex} \\
\fewgen & \LLaMa
& \DeBERTa{} &  69.457895 &  \best{72.6}\ig{15385} &  32.09 &    62.35 &  74.433333 &   80.996 & 65.32
\\


\fewgen & \Claude 
& \DeBERTa{} &  75.031579 &  57.538462 &  23.298 &     47.1 &  49.866667 &   87.481 &  56.72
\vspace{1.0ex}
\\

\ours & \LLaMa 
& \DeBERTa{} &  83.510526 &  69.846154 &  \best{74.4} &  \best{68.9}\ig{41667} &  \best{82.5}\ig{82.453333} &   84.707 & \best{77.32}
\\



\ours & \Claude 
& \DeBERTa{} &  \best{83.9}\ig{47368} &  72.307692 &  71.83 &  66.783333 &  62.113333 &  \best{88.7} \ig{88.736}  & 74.29
\vspace{1ex}
\\


\toprule
\multicolumn{10}{c}{\underline{\textsc{Few-shot}}} \vspace{1ex} \\

\fewgen* & \LLaMa
& \DeBERTa{} &  84.2 &  74.461538 &  \best{73.7} &  68.641667 &  88.38 &   90.899 &  80.05
\\
\fewgen* & \Claude
& \DeBERTa{} &  75.852632 &  58.461538 &  72.224 &  68.833333 &  82.94 &  91.2435 & 74.93
\vspace{1.0ex}
\\

\ours${}^{\dagger}$ & \LLaMa
& \DeBERTa{} &  82.992105 &  78.461538 &  73.262 &  \best{72.4} & \best{90.2} \ig{90.186667} &  91.0015 & \best{81.38}
\\

\ours${}^{\ddagger}$ & \LLaMa
& \DeBERTa{} &  \best{85.2} &  \best{79.1} &  72.826 &  71.941667 &  88.773333 &   88.197 & 81.00
\\


\ours${}^{\dagger}$ & \Claude
& \DeBERTa{} &  83.7394 &  72.307692 &  72.832 &  65.408333 &  83.393333 & \best{91.3} \ig{91.2975} & 78.16 \\

\ours${}^{\ddagger}$ & \Claude
& \DeBERTa{} &  83.710526 &  72.00 &  72.45 &  67.816667 &  76.2066 &    87.87 & 76.68 \\
                                
\bottomrule
\end{tabular}
}
\vspace{-1ex}
\caption{
Test Accuracy ($\uparrow$) after distilling \DeBERTaLarge{} student from \LLaMaFull{} and \ClaudeFull. \Contriever{} was used as the retriever in \ours. We report the average of 5 runs and rerun in cases where ~std.~dev.~$\ge$6\% (indicating one or more models failed to converge). The top half considers zero-shot synthesis and bottom half uses in-context learning, and we \textbf{bold} the best result under each paradigm. Notation: *32-shot; ${}^{\dagger}$3-shot \RetrICL{}; ${}^{\ddagger}$32-shot \NoRetrICL.
}

\vspace{-2ex}
\label{tab:student-deb}
\end{table*}

\insection[]{RQ: How is entity similarity in synthetic data affected by grounding to an in-domain corpus?} 
For the \Category{} task we generate $8$K product reviews and randomly select $8$K \gold{} examples. In \autoref{tab:entity-count}, we measure \textit{entity recall}, and find that the occurrence of \gold{} entities is 100\%-140\% higher in \ours{} than \fewgen. The KL divergence of each entity distribution is also lower. We finally consider the \textit{entity coverage} (unique entities) and \textit{entity density} (total entities). Compared to \gold{}, \fewgen{} tends to produce fewer unique entities (places, events, languages, currencies, etc). Each \fewgen{} example also has a lower density of entities, as visible in \autoref{tab:real-synthetic-examples}. \ours{} coverage and density more closely match \gold. \ig{Similar results are observed on other datasets and other entity types (Appendix).} 

\insection[]{RQ: How distributionally similar are our generated examples and human-written examples?} 
We see from MAUVE scores in \autoref{tab:mauve} that zero-shot generations are quite dissimilar in both approaches compared to few-shot methods. Surprisingly, \ours{} generations are much more similar to human text than \fewgen{}, despite the fact that nothing in our content sourcing strategy explicitly guides \ours{} generations to match the distribution of \gold. 

We thus manually inspect generations and discover an interesting pattern which can be attributed to content sourcing. As shown earlier, and in \autoref{tab:real-synthetic-examples}, the density of entities is higher under \ours{}. \fewgen{} produces generations which obey the prompt, but are very bland and do not include specifics. On the other hand, by obtaining information-rich documents, \ours{} is able to ground the task inversion step in details of the retrieved article/product. We hypothesise that this improves the MAUVE score towards \gold, which is similarly grounded in specifics.

\section{Results: Student distillation}
\label{sec:student}

We have established that \ours{} generates more diverse datasets compared to a baseline approach. Now, we return to the application of training a specialist model based on these datasets. 

\autoref{tab:student-deb} shows the results of training a \DeBERTaLarge{} student on datasets generated by \ours{} and \fewgen, as well as baselines of tuning on the \gold{} set and \textsc{Seed} set. In the zero-shot setting, we find that \ours{} performs much better than \fewgen, despite using the same frozen teacher LLM. Note that \ours{} uses in-context examples for retrieval here whereas \fewgen{} does not; our method has some additional supervision here. However, in this setting, we see clear gains during the task inversion stage ($\uparrow$12\% for \LLaMaShortNonSc{} and $\uparrow$17.6\% for \ClaudeShortNonSc). Thus, having access to retrieval yields a better final dataset, almost on par with 32-shot \fewgen.

With ICL, 3-shot \ours{} using the \RetrICL{} strategy trains better students than 32-shot \fewgen{} ($\uparrow$1.3\% for \LLaMaShortNonSc{} and $\uparrow$3.2\% for \ClaudeShortNonSc) and \NoRetrICL. We conclude that naively adding ICL examples is not an effective use of the LLM's context window. Instead, a better content sourcing strategy improves the student distillation, leading to better test performance.

\begin{table*}[!t]
\centering
\sisetup{round-mode=places, round-precision=1} 
\setlength{\tabcolsep}{1.9pt}
\footnotesize{

\begin{tabular}{
C{1.6cm}  
C{1.15cm}  
C{1.8cm}  
h 
c 
C{0.7cm} 
c 
h 
|c 
C{0.7cm} 
c  
h 
|c 
C{0.7cm} 
c 
h 
|c 
C{0.7cm} 
c 
}
\toprule
\textbf{Method}
& \multirow{2}{*}{\textbf{Retriever}}
& \textbf{Teacher}
& \multicolumn{4}{c}{\underline{Self-BLEU-5 \lowerbetter}} 
& \multicolumn{4}{c}{\underline{Entity Entropy \higherbetter}} 
& \multicolumn{4}{c}{\underline{Mauve \higherbetter}} 
& \multicolumn{4}{c}{\underline{Accuracy \higherbetter}} 
\\
[0.7ex]
\textit{(Dataset)}
& 
& \textbf{LLM}
& \Yelp 
& { }\AG 
& \IMDb 
& { }\SST 
& { }\Yelp 
& { }\AG 
& \IMDb 
& { }\SST 
& \Yelp 
& { }\AG 
& \IMDb 
& { }\SST 
& \Yelp 
& { }\AG 
& \IMDb 
& { }\SST 
\\ 
\midrule
 \gold &  - & - 
 & 27.5  &  17.1  & 27.9   &  35.5  
 & 7.1   & 6.6   &  7.5  &  3.2 
 &  -  &  -  &  -  & -   
 & 94.5  & 90.8   & 91.3   & 88.2
 \vspace{0.75ex} \\ 
\toprule 
\\
 \SunGen \ig{gao2023selfguided}  & - &  \GPTTwoXL  
 &  \norelease{}  &  \norelease{}  &  \textbf{15.4}  &  \norelease{}  
 &  \norelease{}  &  \norelease{}  &  4.9  &  \norelease{}  
 &  \norelease{}  &  \norelease{}  &  68.7  &  \norelease{}  
 & \norelease{}  &  \norelease{}  &  84.9  &  \norelease{}
 \\
 \ReGen \ig{yu-etal-2023-regen} &  \BERT & -   
 &  \norelease{}  &  56.5  &  \norelease{}  &  \norelease{}  
 &  \norelease{}  &  \textbf{8.1}  &  \norelease{}  &  \norelease{}  
 &  \norelease{}  &  68.1  &  \norelease{}  &  \norelease{}  
 &  \norelease{}  &  82.7  &  \norelease{}  & \norelease{}
 \\
 \LetsSynth \ig{wang-etal-2023-lets} & - &  \GPTDavinci  
 &  \noeval  &  \noeval  &  62.2  &  \noeval  
 &  \noeval  &  \noeval  &  5.7  &  \noeval  
 &  \noeval  &  \noeval  &  62.0  &  \noeval  
 &  \noeval  &  \noeval  &  \textbf{87.1}  &  \noeval
 \\
 \AttrPromptShort \ig{yu2023large} & - &  \ChatGPTShort  
 &  70.3  &  39.8  &  \norelease  &  71.5  
 &  3.4  &  6.0  &  \norelease  &  3.4  
 &  55.9  &  52.8  &  \norelease  &  50.0  
 &  89.3  &  79.8 & \norelease{}  &  80.8
 \\
 \toprule
\multicolumn{19}{c}{\underline{\textsc{Zero-shot}}} \vspace{1ex} \\
 \oursshort & \ContrieverShort  &  \LLaMa 
 &  70.2  &  29.3  &  66.3  &  41.9  
 &  5.4  &  7.1  &  5.7  &  4.5  
 &  67.9  &  89.5  &  58.5  &  50.0  
 &  85.6  &  85.3  &  82.9  &  80.2
 \\
 \oursshort & \ContrieverShort  &  \Claude 
 &  70.3  &  31.5  &  51.5  &  45.3  
 &  \textbf{5.5}  &  6.6  &  5.3  &  4.8  
 &  54.0  &  94.2  &  55.9  &  50.0  
 &  87.5  &  85.6  &  83.6  &  82.5
 \\
 [0.75ex]
 \oursshort & \BM  &  \LLaMa 
& 70.6  &  28.7  & 62.2   &  36.5  
 & 4.7   & 7.0   &  5.6  &  5.1
 & 64.7   &  90.3  &  60.5  & 50.0  
 & 83.2  & 84.3   & 74.1   &  \best{84.4}
 \\
 \oursshort & \BM  &  \Claude 
 & 72.9  &  30.9  & 50.4   &  36.9  
 & 4.5   & 6.5   &  5.1  &  \best{5.4}
 &  54.6  &  90.8  &  53.2  & 50.0  
 & 86.7  & 84.2   & 79.1   & 82.6
 \\
[0.75ex]
\multicolumn{19}{c}{\underline{\textsc{3-shot \RetrICL}}} \vspace{1ex} \\
 \oursshort & \ContrieverShort &  \LLaMa  
 &  59.5  &  34.2  &  62.9  &  26.3  
 &  4.9  &  7.2  &  5.7  &  3.8  
 &  \textbf{84.9}  &  92.6  &  72.6  &  50.0  
 &  89.9  &  84.6  &  84.8  &  83.8 
 \\
 \oursshort & \ContrieverShort &  \Claude  
 &  \textbf{58.5}  &  \textbf{23.7}  &  38.0  &  \textbf{24.6} 
 &  4.9  &  6.7  &  \textbf{5.9}  &  4.3 
 &  70.3  &  95.8  &  58.0  &  50.0  
 &  \textbf{91.3}  &  \textbf{86.0}  &  86.3  &  80.6
 \\
 [0.75ex]
 \oursshort & \BM  &  \LLaMa 
 & 66.6  &  32.0  & 59.7   &  25.3  
 & 5.1   & 7.2   &  5.6  &  4.8
 &  81.9  &  92.5  &  \best{78.7}  & 50.0  
 &  90.7 & 84.3   & 84.7  & \best{84.4}
 \\
 \oursshort & \BM  &  \Claude 
 & 59.9  &  24.6  & 41.9   &  26.8  
 & 5.0   & 6.7   &  5.4  &  4.9 
 &  63.3  &  \best{96.0}  &  58.5  & 50.0  
 & 89.4  & 84.1   & 81.6  & 82.3
 \\
\bottomrule
\end{tabular}
}
\vspace{-1ex}
\caption{
Evaluations of synthetic datasets released by prior work. We subsample all to 6K examples (uniformly distributed across classes) before computing metrics as described in \S\ref{sec:expt-setup}. Tasks not evaluated by previous authors are denoted by {\noeval} while those evaluated without dataset release are marked {\norelease}. \GPTDavinci{} is text-davinci-003 whereas \ChatGPTShort{} is gpt-3.5-turbo \citep{openai2022gpt35}, \LLaMa{} is 13B Chat version \citep{touvron2023llama}, \Claude{} is Instant-V1.2 version \citep{anthropic2023claudev12}. Accuracy is measured on a \DistilBERT{} student, where we train 5 student models and report the mean accuracy (std.~dev.~was $\le 2.0$ in all cases). Within each dataset, we \textbf{bold} the best result.
}
\vspace{-3ex}
\label{tab:baselines-combined-detailed}
\end{table*}

\section{Comparison to previous work}
\label{sec:detailed-baseline-comparison}


We benchmark \ours{} against four prior synthesis methods: 
(1) \textbf{\SunGen{}} \cite{gao2023selfguided} uses \ZeroGen{} to create 200k synthetic rows and employs a custom bi-level optimization algorithm to weight each instance; 
(2) \textbf{\ReGen{}} \cite{yu-etal-2023-regen} utilizes two BERT models, one for retrieval and one as a classifier, to multi-round filter noisy data; 
(3) \textbf{\LetsSynth{}} \cite{wang-etal-2023-lets} builds and iteratively enhances a seed dataset by identifying and synthesizing corrections using an LLM; 
(4) \textbf{\AttrPrompt{}} \cite{yu2023large} improves dataset diversity and unbiasedness by prompting \ChatGPT{} with varied attributes (derived from a human-in-the-loop analysis of each task). Standard zero-shot and few-shot generation baselines were compared in \autoref{tab:student-deb}, so we do not include them here. \ZeroGen{} \citep{Ye2022ZeroGenEZ} is similarly excluded.



We benchmark three popular tasks: \IMDb{} \cite{maas-etal-2011-learning}, \SST{} \cite{socher-etal-2013-recursive} and \AGNews{} \citep{zhang2015character}. 
Previous studies have generated larger datasets ranging from 20k to 200k examples with varying student model hyperparameters, but often lack reports on intrinsic dataset quality, making a fair comparison challenging. Therefore, we independently reproduce these results using the synthetic datasets released by the original authors\footnote{\ProGen{} \citep{ye-etal-2022-progen} was excluded as it does not release datasets.}. Following \citet{yu2023large}, we subsample these datasets to 6k rows, keeping a uniform distribution across classes, and generate the same number of synthetic covariates using \ours{} \RetrICL{} (Algorithm~\ref{alg:synthesizrr}). For the content sourcing stage of \ours, we retrieve documents from the \CMUMovies\ corpus \cite{bamman-etal-2013-learning} and \RealNewsDom (\autoref{sec:preprocessing}). We measure accuracy on a \DistilBERT{} student \citep{Sanh2019DistilBERT, yu2023large, Ye2022ZeroGenEZ, gao2023selfguided, wang-etal-2023-lets, ye-etal-2022-progen}, fixing hyperparams to \citet{yu2023large}. 

\insection[]{RQ: How does \ours{} perform against prior methods on student model accuracy?}

Methods like \SunGen{} rely on relatively weak LLM teachers like \GPTTwoXL{} \cite{radford2019language} can perform well on topic and sentiment tasks like \IMDb, but require a very high data cost (15-30x more synthetic data than \ours). In \autoref{tab:baselines-combined-detailed}, we observe that when scaled down to 6k rows, the performance deteriorates significantly. We hypothesize that adding the student model into the synthesis process impacts the final classification accuracy, as the dataset becomes specialized to the particular choice of student and does not generalize to other students.

Approaches which use strong instruction-following LLMs like \AttrPrompt, \LetsSynth, and \ours{} can achieve similar or better performance with much smaller datasets, as they create high-quality datasets. Prompting techniques like Chain-of-Thought \citep{wei2022chain} used by \LetsSynth{} further improve the task-inversion step (while necessitating higher API costs due to longer output lengths). Chain-of-Thought prompting thus seems like a promising approach to augment \ours{}'s task-inversion step.


\insection[]{RQ: do we find evidence that content sourcing promotes diversity and similarity?}

\autoref{tab:baselines-combined-detailed} compares diversity (Entity-Entropy, Self-BLEU), and similarity to \gold{} texts (MAUVE). Only \AttrPrompt{} \citep[Appendix E]{yu2023large} attempts to improve diversity of the generated text, by templatizing the task inversion instruction with attributes such as \paramnorm{style, topic, length:min-words} and more. \ReGen{} is the only prior approach to use content sourcing (but not task inversion). These are thus the most relevant baselines for \ours.

Both \ReGen{} and \ours{} achieve very high entity entropy compared to \AttrPrompt, underscoring the importance of a content sourcing step. Unlike \ours, \ReGen{} uses only retrieval without task-inversion, and thus suffers in terms of lexical diversity, MAUVE and student accuracy. 

On the other hand, CoT-style prompting (\LetsSynth) suffers a lack of lexical diversity and similarity to \gold{} texts, despite strong distillation performance. This is reproduced in \AttrPrompt{} and previously in \fewgen, lending evidence to our claim that synthesis without content sourcing tends to produce datasets with lower diversity, which cannot be overcome by complex prompting strategies alone.


Finally, \SunGen{} exhibits high diversity on \IMDb, a task for generating sentiment-based movie reviews. Unlike traditional zero-shot generation, \SunGen{} begins by creating a movie with the prompt \paramnorm{Movie:} followed by generating an example using prompt \paramnorm{The movie review in positive sentiment for movie "<Movie>" is:} (details in \citet[Section~4.6]{Ye2022ZeroGenEZ}). We posit that this generated movie fulfils a similar purpose to a retrieved context, enhancing the diversity.

\section{Related Work}
\vspace{-0.5em}

\insection{Dataset synthesis using LLMs} Using LLMs to perform \textit{task inversion} for dataset synthesis has been studied previously. Most use GPT-2XL without fine-tuning \citep{ye-etal-2022-progen, Ye2022ZeroGenEZ, gao2023selfguided, meng2022generating, schick-schutze-2021-generating, jung2023impossible}. Recent work has considered large teacher LLMs such as GPT-3 \citep{west-etal-2022-symbolic, honovich-etal-2023-unnatural, wang-etal-2023-self-instruct}, PaLM-540B \citep{hsieh-etal-2023-distilling} and chat-tuned LLMs such as \mbox{gpt-3.5-turbo} \citep{yu2023large, yehudai2024genie, wang-etal-2023-lets}. 

For the generation of text classification datasets, class-conditioned prompting is key. Prior approaches investigated zero-shot \citep{Ye2022ZeroGenEZ} and iterative few-shot prompting \citep{ye-etal-2022-progen}, or synthesis using seq2seq LLMs fine-tuned on a curated dataset \citep{Lee2021NeuralDA}. Recently, \AttrPrompt{} \citep{yu2023large} established that varying prompt attributes improves diversity. 
Our work explores adding retrieval contexts as the source of diversity. 

\insection{Retrieval-augmented generation} 
Our approach has many of the characteristics of in-context retrieval-augmented generation (RAG) \citep{lewis2020retrievalaugmented,ram-etal-2023-context, huang2023raven, atlas}. Previous studies show how RAG bypasses numerous problems associated with generating solely from parametric memory, i.e., heightened bias towards ``head'' entities \citep{mallen-etal-2023-trust}, lower lexical diversity  \citep{Holtzman2020CuriousCase, jentzsch-kersting-2023-chatgpt}, and hallucinated information \citep{zhang2023hallucination}.

Using retrieval-augmented generation for synthesis of classification tasks has not been explored at the instance level. \ReGen{} \citep{yu-etal-2023-regen} studies the retrieval-only setting for creation of topic and sentiment datasets, which are simpler than the tasks in our work. 
\citet{viswanathan-etal-2023-prompt2model} and \citet{gandhi2024better} perform dataset-level retrieval and not instance-level retrieval.
\section{Conclusion}
\vspace{-0.5em}

In this work we describe how a retrieval corpus can be used to aid the synthesis of a text classification data set in specialized domains. We show that the diversity of the generated data is enhanced by including retrieved documents in a generation prompt. Compared to few-shot generation, we find that \ours{} produces more diverse and representative text and leads to better students. 
\section*{Limitations}

Most principally, our work relies on the existence of a large corpus that is close enough to the task at hand. This may be prohibitive for doing dataset generation in low-resource languages, where a large corpus of related content may not be available. It would be intriguing to explore cross-lingual transfer of content sourcing, but this would require additional experimental validation. By contrast, approaches like \textsc{FewGen} do not require this corpus.

The need for an explicit context sourcing step and increased prompt-length causes an increase in the expenses and latency, especially when using LLM APIs. Such increased expense may not be worth it in the presence of a poor quality retrieval corpus. For one, if the in-context examples are not easily reusable as queries, then \ours{} can retrieve irrelevant documents which might not be suitable for task inversion. Furthermore, in the case of factually dubious corpus documents, the student model may end up grounding in factually incorrect information. This can be mitigated by a human-in-the-loop step to remove such documents before task inversion.

Finally, we note that the scope of our experiments is restricted to a set of classification tasks over a few English domains of text. While we believe our approach can be applied to other languages, other domains, and tasks like question answering that go beyond classification, we have not validated this in this work.



\appendix

\section{Risks}

Although the main goal of our work is to improve text classification, our use of LLMs to generate examples does carry some conceptual risks. By generating news articles to train classifiers on, we run the risk of generating fake news and other harmful content. However, we believe this risk is mitigated by the fact that the final outcome of our system is a classifier: classification models have relatively constrained failure modes (misclassification) compared to text generation models that can mislead users. Furthermore, we do not believe our approach uniquely advances the generation of content like fake news; our advances are largely orthogonal to the technology that brings such risks.

\section{Incorporating feedback from distilled student models}
\label{sec:student-analysis}

\insection[]{RQ: Why does \ours{} improve classification dataset synthesis?} In this section we take a closer look at the generated classification dataset and how it affects the \textit{training dynamics} of student models during distillation. 

Aside from the final accuracy, we also consider \textbf{label preservation accuracy}, which is obtained from an ``oracle'' model for the task. We construct this oracle from \gold{} data by running a grid-search over \DeBERTaLarge{} hyperparams (\autoref{sec:hyperparams}), splitting 80\% of the \gold{} train set for fine-tuning and 20\% for validation. Then, we measure the fraction of synthetic examples which the oracle classifies to belong to the prompted target class. This indicates the adherence of the generated example to the class it \textit{should} belong to, as per the prompt. 

We would expect that better label preservation means a higher-fidelity training dataset. However, \autoref{tab:label-pres} shows that \fewgen{} datasets have very high label preservation in spite of their lower test performance. Especially on multiclass tasks (\AG, \ToI, \Cat), \fewgen{} shows the highest label preservation (exceeding \gold) but this does not translate into improved student performance. 

\begin{figure}[!t]
\includegraphics[width=0.47\textwidth]{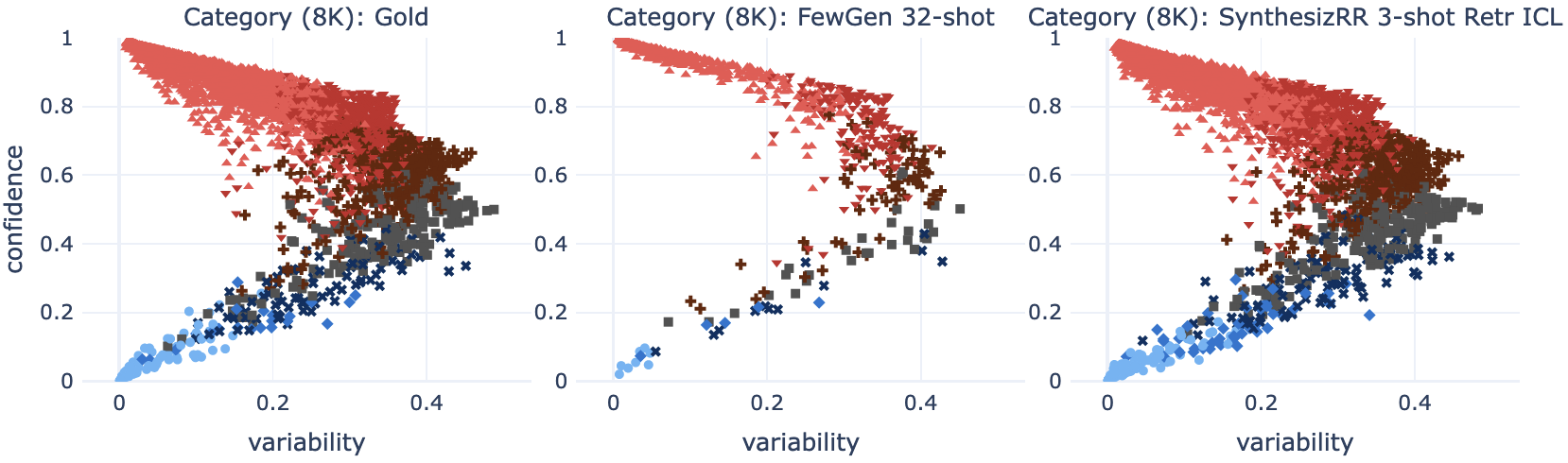}

\centering
\vspace{-0.5ex}
\caption{
Data maps from a \DistilBERT{} training run on $8$K \Category{} rows from \LLaMa. \fewgen{} (center) is skewed towards easy-to-learn examples (top-left) while \gold{} (left) and \ours{} (right) have a higher density of ambiguous examples.
}
\vspace{-3ex}
\label{fig:data-map-category}
\end{figure}

To understand this, we conduct a deeper analysis of the student training dynamics on multiclass datasets. We train a \DistilBERT{} student for 6 epochs and plot the corresponding data-maps \citet{swayamdipta-etal-2020-dataset}. For binary tasks, the data-maps for \ours{} matched both \fewgen{} and \gold, but the data maps from multi-class differed greatly. \autoref{fig:data-map-category} illustrates this difference using the \Category{} task maps. From \autoref{fig:data-map-category} it is clear that \fewgen{} generations tend to cluster around easy-to-learn examples (high confidence and low variability), whereas \ours{} contains more ambiguous examples (high variability) which \citet{swayamdipta-etal-2020-dataset} demonstrate is essential to learning the nuances between classes. 

\insection[]{RQ: Can we improve distillation performance by leveraging student feedback from data-maps?}

\citet{swayamdipta-etal-2020-dataset} use data-maps to filter out easy to-learn examples (top-left, red) and potentially mislabelled examples (bottom-left, blue) and obtain superior accuracy on human-generated datasets. We attempt to apply this same technique to the synthetic datasets generated by \ours{} and \fewgen. 

Concretely, we filter out the least ambiguous examples (bottom 17\% variability) and retrain the \DistilBERT{} student model on the smaller, filtered dataset. In \autoref{tab:student-cart} we find that \fewgen{} performance degrades, whereas \ours{} improves (giving us new best performances on multi-class despite using only 83\% of rows). We conclude that \ours{} generates more ambiguous examples, and this helps establish better class-separability in multi-class data sets.

\begin{table}[!t]
\sisetup{round-mode=places, round-precision=1} 
\centering
\setlength{\tabcolsep}{1pt}
\footnotesize{
\begin{tabular}{
l
S[table-format=3.2]
S[table-format=3.2]
S[table-format=3.2]
S[table-format=3.2]
S[table-format=3.2]
S[table-format=3.2]
}
\toprule
\textbf{Method}
& \AG           & \Hyp        & \ToI         &  \Cat     & \Hum         & \Pol \\ 
\textit{(Dataset size)} & \text{($8$K)}    & \text{($2$K)} & \text{($8$K)}  &  \text{($8$K)}   & \text{($2$K)}  & \text{($4$K)} \\ 
\midrule 
\gold & 93.81 &  81.55 &  85.23 &     84.8 &  95.5 &     96.6     \\

\toprule




\multicolumn{7}{c}{\underline{\LLaMa{} \textsc{Few shot}}} \vspace{1ex} \\
\fewgen*                  & \best{92.4} &   71.25 &   \best{85.9} &   \best{88.1} &    71.7 &    94.75     \\
\oursshort${}^{\dagger}$ & 86.88 &   \best{78.6} &        74.29 &        72.11 &        90.65 &        94.77     \\
\oursshort${}^{\ddagger}$  & 87.64 &         75.5 &        74.85 &  74.46 & \best{95.7} & \best{97.6}     \\

\multicolumn{7}{c}{\underline{\Claude{} \textsc{Few shot}}} \vspace{1ex} \\
\fewgen*                & \best{94.5} &   63.75   & \best{87.4} &  \best{89.4} &   85.85 &    99.55 \\
\oursshort${}^{\dagger}$ &  87.56   &  \best{72.8} & 74.79 &        69.41 &         \best{90.7} &        99.32 \\
\oursshort${}^{\ddagger}$  &  87.36 &        65.85 & 73.24 &        73.24 &        77.35 &         \best{99.7} \\

\bottomrule

\end{tabular}
}
\vspace{-1ex}
\caption{
Few-shot label-preservation accuracy ($\uparrow$) using tuned oracle \DeBERTa{} model. \gold{} row is accuracy on 20\% validation split. Notation: *32-shot; ${}^{\dagger}$3-shot \RetrICL{}; ${}^{\ddagger}$32-shot \NoRetrICL.
}
\label{tab:label-pres}
\end{table}

\begin{table}[!t]
\sisetup{round-mode=places, round-precision=1} 
\centering
\setlength{\tabcolsep}{3pt}
\footnotesize{
\begin{tabular}{
lh
C{11pt}c
C{11pt}c
C{11pt}c
C{28pt}
}
\toprule
\textbf{Method}
& \multirow{2}{*}{\textbf{LLM}}  
& 
\multicolumn{2}{c}{\AG}  & \multicolumn{2}{c}{\ToI} & \multicolumn{2}{c}{\Cat} 
& \multirow{2}{*}{\textbf{Avg}}
\\ 
\textit{(Dataset size)} 
&
& \multicolumn{2}{c}{\text{($6.6$K)}}  & \multicolumn{2}{c}{\text{($6.6$K)}} & \multicolumn{2}{c}{\text{($6.6$K)}} &
\\ 
\midrule 

\multicolumn{9}{c}{\underline{\LLaMa{} \textsc{Few shot}}} \vspace{1ex} \\
\fewgen* & \LLaMa
&  58.0\ig{6.2}    & \redtext{${\downarrow}$26.2}
&  37.6\ig{3.8} & \redtext{${\downarrow}$36.1}
&  48.0\ig{2.7}  & \redtext{${\downarrow}$20.6}
& \redtext{${\downarrow}$27.6}
\\
\oursshort${}^{\dagger}$ & \LLaMa
& 85.7 \ig{0.4} & \darkgreentext{${\uparrow}$2.7}
& 76.0\ig{0.9}  & \darkgreentext{${\uparrow}$2.7}
& 74.3\ig{1.0} & \darkgreentext{${\uparrow}$1.9}
& \darkgreentext{${\uparrow}$2.4}
\\
\oursshort${}^{\ddagger}$  & \LLaMa
& 86.3\ig{1.3}      & \darkgreentext{${\uparrow}$1.1}
& 75.0\ig{1.6}  & \darkgreentext{${\uparrow}$2.2}
& 72.9\ig{0.7}      & \darkgreentext{${\uparrow}$1.0}
& \darkgreentext{${\uparrow}$1.4}
\\

\multicolumn{9}{c}{\underline{\Claude{} \textsc{Few shot}}} \vspace{1ex} \\
\fewgen* & \Claude
& 71.8\ig{1.1} & \redtext{$\downarrow$4.1}
& 72.1\ig{2.3}      & \redtext{$\downarrow$0.1}
& 69.3\ig{2.1} & \darkgreentext{$\uparrow$0.5}
& \redtext{${\downarrow}$1.2}
\\
\oursshort${}^{\dagger}$ & \Claude
& 86.2\ig{1.0}  & \darkgreentext{$\uparrow$2.5}
& 75.3\ig{0.5}      & \darkgreentext{$\uparrow$2.5}
& 69.0\ig{1.8}      & \darkgreentext{$\uparrow$3.6}
& \darkgreentext{${\uparrow}$2.9}
\\
\oursshort${}^{\ddagger}$  & \Claude
& 86.1 \ig{0.5}  & \darkgreentext{$\uparrow$2.4}
& 74.6\ig{1.3} & \darkgreentext{$\uparrow$2.1}
& 70.0\ig{1.4}  & \darkgreentext{$\uparrow$2.2}
& \darkgreentext{${\uparrow}$2.2}
\\

\bottomrule

\end{tabular}
}
\vspace{-1ex}
\caption{
Test Accuracy \higherbetter{} after keeping 83\% most-ambiguous examples. We report improvements compared to \autoref{tab:student-deb}. Notation: *32-shot; ${}^{\dagger}$3-shot \RetrICL{}; ${}^{\ddagger}$32-shot \NoRetrICL.
}
\vspace{-3ex}
\label{tab:student-cart}
\end{table}

\section{Bootstrapping with a synthetic seed set}
\label{sec:bootstrap-seed}

A core assumption in \ours{} has been the existence of a small seed set of human-written $(x, y)$ pairs for the task. This seed set is critical as it serves a dual purpose: it is used as the set of the retrieval queries, and as in-context learning examples to guide the teacher LLM's next-token distribution in the task inversion step.

In this section we consider how we can synthesize such a seed set for low-resource settings. Our core assumption is that the seed set is small (100/class for binary tasks and 50/class for multiclass tasks). Thus using \fewgen{} with top-$p=0.9$ and temperature $=0.95$ and three in-context examples, we attempt to generate a diverse seed set with minimal repetitions. This bootstrapping approach makes \ours{} tractable when very little human data is available (just 5-15 examples per class) or no human data is available. 

Concretely, we compare three paradigms: 

\begin{enumerate}
\item \textbf{True zero-shot:} when we have no human data we utilize zero-shot generation to bootstrap the seed set.
\item \textbf{Low-resource:} Here, we assume we have a small number of human-written examples, e.g. 5 examples per class. This is presumed insufficient to be used as the seed set directly, but we can use it as in-context examples to guide the \fewgen{} generator to bootstrap a realistic seed set. 
\item \textbf{Sufficient:} We do not synthesize the seed set. This is the \ours{} paradigm we have explored in previous sections, wherein we have 50-100 \gold{} examples per class in our seed set. 
\end{enumerate}

As mentioned in \S\ref{sec:expt-setup}, the true zero-shot paradigm makes strong assumptions that are often unnecessarily restrictive. In practice, it is typically feasible to obtain a small amount of human-written examples (low-resource or sufficient seed), while obtaining several thousand human-written examples is still challenging.

\begin{table}[!t]
\centering
\sisetup{round-mode=places, round-precision=1} 
\setlength{\tabcolsep}{2pt}
\footnotesize{

\begin{tabular}{
C{1.2cm}  
C{1.3cm}  
C{0.65cm} 
C{0.65cm} 
C{0.65cm} 
C{0.65cm} 
C{0.65cm} 
C{0.65cm} 
h  
}
\toprule
\textbf{\gold}
& \textbf{\RetrICL}
& \AG          
& \Hyp        
& \ToI         
& \Cat     
& \Hum         
& \Pol 
& { }\multirow{2}{*}{\textbf{Avg.}}
\\
\textbf{data ($N$)}
& \textbf{shots}
& \text{($8$K)}    
& \text{($2$K)} 
& \text{($8$K)}  
& \text{($8$K)}   
& \text{($2$K)}  
& \text{($4$K)} 
&
\\ 
\midrule
\multicolumn{9}{c}{\underline{\textsc{\gold}}} 
\vspace{1ex} \\
All  &  -  &  91.0 &  93.2 &  82.5 &  81.5 &  93.1 &  95.3  & 89.43 \\
[1.0ex]
\toprule
\multicolumn{9}{c}{\underline{\textsc{True Zero-shot (0-shot \fewgen{} seed) }}} 
\vspace{1ex} \\
None  & 0-shot & \textbf{66.6}  & 68.0  & 60.5   & 60.4   & \textbf{76.9} & 76.4 & N/A \\
[0.5ex]
None  & 3-shot & 60.0  & \textbf{72.3}  & \textbf{62.5}  & \textbf{61.7}  & 72.3 & \textbf{85.4} & N/A \\
[1.0ex]
\toprule
\multicolumn{9}{c}{\underline{\textsc{Low-Resource ($\binom{N}{3}$-shot \fewgen{} seed) }}} 
\vspace{1ex} \\
5/class  & 0-shot & \textbf{79.9}  & 71.7  & 68.1  & 63.4  & 81.3 & 81.3 & N/A \\
[0.5ex]
5/class  & 3-shot & 77.7 &  66.8 & 68.9   & 58.8  & \textbf{86.4}  & \textbf{86.5} & N/A \\
[1.0ex]
15/class  & 0-shot & 78.5  & \textbf{72.9}  & \textbf{69.3}   & \textbf{65.7} & 77.4  & 84.0 & N/A \\
[0.5ex]
15/class  & 3-shot & 76.1  & 72.6  & 71.6  & 63.5  & 82.5 & 73.8 & N/A \\
[1.0ex]
\toprule
\multicolumn{9}{c}{\underline{\textsc{Sufficient (\gold{} Seed)}}} 
\vspace{1ex} \\
\mbox{Full seed}  & 0-shot  & \best{83.5} & 69.8 &  \best{74.5} &  68.9 & 82.5 &  84.7 & \best{77.32} \\
[0.5ex]
\mbox{Full seed}  & 3-shot  & 83.0 &  \best{78.5} &  73.3 &  \best{72.4} & \best{90.2} &  \best{91.0} & \best{81.38} \\
\bottomrule
\end{tabular}
}
\vspace{-1ex}
\caption{
Test accuracy after distilling a \DeBERTa{} student on a dataset generated by \ours{} \RetrICL{} variant. We use the same corpus as \autoref{tab:corpus}, but vary the seed set. \LLaMaFullNonSc{} is used as the teacher LLM. We train 5 student models and report the mean accuracy, rerunning all 5 in case of std $\ge$ 6.0. ``'Full'' seed implies 100 \gold{} examples per class for binary and 50 per class for multiclass tasks. We \textbf{bold} the best result in each paradigm.
}
\vspace{-3ex}
\label{tab:bootstrap-seed}
\end{table}

The results of running \ours{} \RetrICL{} using synthetic seed data is shown in \autoref{tab:bootstrap-seed}. As a general trend, adding more human-written examples leads to better performance. Unsurprisingly, the best results are in the Sufficient paradigm, where we use 50-100 \gold{} examples as both retrieval queries and the the \RetrICL{} set. True Zero-shot results (without any human input) are considerably worse. Surprisingly, however, we are able to get good distillation accuracy with just 5 examples per class rather than the full 50-100 per class, which indicates that \ours{} might be usable in low-resource settings where human annotated data is scarce. 

In certain cases of the low-resource paradigm, we observe that the performance drops significantly from 0-shot \RetrICL{} to 3-shot \RetrICL. We attribute this to the fact that, even with 5-15 \gold{} in-context examples, the \fewgen-generated seed set might not be reflective of the true \gold{} examples (this behavior is reflected in the low MAUVE scores in \autoref{tab:mauve}). Thus, by conditioning on \mbox{incorrect} synthetic examples during \RetrICL, we shift the next-token distribution away from the true distribution.

In conclusion, using \fewgen{} to bootstrap a seed set can be a viable approach to using \ours{} in low-resource settings where there is not enough \gold{} task-data. 
\section{Influence of corpus on domain shift}
\label{sec:domain-shift}
\begin{table}[!t]
\centering
\footnotesize{
\setlength{\tabcolsep}{2pt}

\begin{tabular}{
L{38pt}
h
C{42pt}
h
h
C{28pt}
h
h
h
h
C{50pt}
h
C{36pt}
}
\toprule
\multicolumn{13}{c}{\underline{\AGNews{} ($4$K)}} \vspace{0.5ex} \\
Corpus & Dataset & \textsc{DeBERTa} ($\uparrow$) & \DistilBERT{} ($\uparrow$) &   \TinyBERT{} ($\uparrow$) &     Mauve ($\uparrow$) &  Self-BLEU-1 ($\downarrow$) &  Self-BLEU-2 ($\downarrow$) &  Self-BLEU-3 ($\downarrow$) &  Self-BLEU-4 ($\downarrow$) &  \mbox{Self-BLEU-5} ($\downarrow$) & Label Acc. ($\uparrow$) & \mbox{Entity Ent.} ($\uparrow$) \\
\midrule

\RNDom        &        \AGNews{} (4K) &       \best{85.39 ± 0.8} &  85.12 ± 0.3 &  82.17 ± 0.5 & \best{92.58} &         0.96 &         0.78 &         0.55 &         0.36 &         0.23 &      86.4\% & 6.72 \\
\textsc{RN/Rnd} &        \AGNews{} (4K) &       35.57 ± 6.1 &  40.53 ± 1.0 &  45.39 ± 2.9 &  83.39 &         0.94 &         0.74 &         0.50 &         0.33 &         \best{0.22} &      40.9\% & \best{7.07} \\
\RNReg        &        \AGNews{} (4K) &       84.17 ± 0.7 &  81.66 ± 0.4 &  79.18 ± 0.7 &  88.88 &         0.96 &         0.79 &         0.57 &         0.39 &         0.26 &      82.6\%  & 6.72 \\
\midrule

\multicolumn{13}{c}{\underline{\Hyperpartisan{} ($2$K)}} \vspace{0.5ex} \\
Corpus & Dataset & \textsc{DeBERTa} ($\uparrow$) & \DistilBERT{} ($\uparrow$) &   \TinyBERT{} ($\uparrow$) &     Mauve ($\uparrow$) &  Self-BLEU-1 ($\downarrow$) &  Self-BLEU-2 ($\downarrow$) &  Self-BLEU-3 ($\downarrow$) &  Self-BLEU-4 ($\downarrow$) &  \mbox{Self-BLEU-5} ($\downarrow$) & Label Acc. ($\uparrow$) & \mbox{Entity Ent.} ($\uparrow$) \\
\midrule

\RNDom        &  \Hyperpartisan{} (2K) &       \best{78.77 ± 2.8} &  65.85 ± 5.8 &  64.62 ± 4.5 &  \best{66.94} &         0.96 &         0.83 &         0.65 &         0.48 &         0.35 &      80.3\% & 6.11 \\
\textsc{RN/Rnd} &  \Hyperpartisan{} (2K) &       78.77 ± 3.5 &  67.08 ± 4.2 &  63.38 ± 5.7 &  61.45 &         0.93 &         0.76 &         0.55 &         0.38 &         \best{0.25} &      69.3\% & \best{7.40} \\
\RNReg        &  \Hyperpartisan{} (2K) &       72.00 ± 2.0 &  62.46 ± 4.8 &  67.08 ± 1.4 &  65.59 &         0.95 &         0.83 &         0.65 &         0.49 &         0.35 &      78.6\% & 6.12 \\

\bottomrule
\end{tabular}
\vspace{-1ex}
}
\caption{
Effect of corpus-swapping for \ours{} \mbox{32-shot} \NoRetrICL. We generate only $4$k rows for \AGNews{} to reduce costs.
}
\vspace{-1ex}
\label{tab:domain-shift}
\end{table}

Our expectation is that \ours{} can flexibly specialize students to different domains by transparently changing the retrieval corpus, while keeping a frozen LLM. To quantify how changing the retrieval corpus might affect earlier metrics, we switch the news corpus for \Hyperpartisan{} and \AGNews{}. We had assumed \RealNewsDom{} was the most suitable corpus (in-domain), and the others will cause domain-shift. In the following RQs, we validate the degree to which this assumption holds and the importance of information retrieval as the content sourcing mechanism in \ours.

\insection[]{RQ: Does modifying the corpus cause domain shift?}
\autoref{tab:domain-shift} finds that the retrieval corpus highly influences the test performance (both student and intrinsic metrics). When grounding to a corpus with highly dissimilar entities (such as \RealNewsReg), all metrics drop significantly. Thus, we can conclude that an alternative content-source does indeed induce domain-shift. Mauve and distillation accuracy are highest for the in-domain corpus, while Self-BLEU and Entity entropy are highest for the random-retrieval results. 



\begin{figure}[t!]
\includegraphics[width=0.47\textwidth]{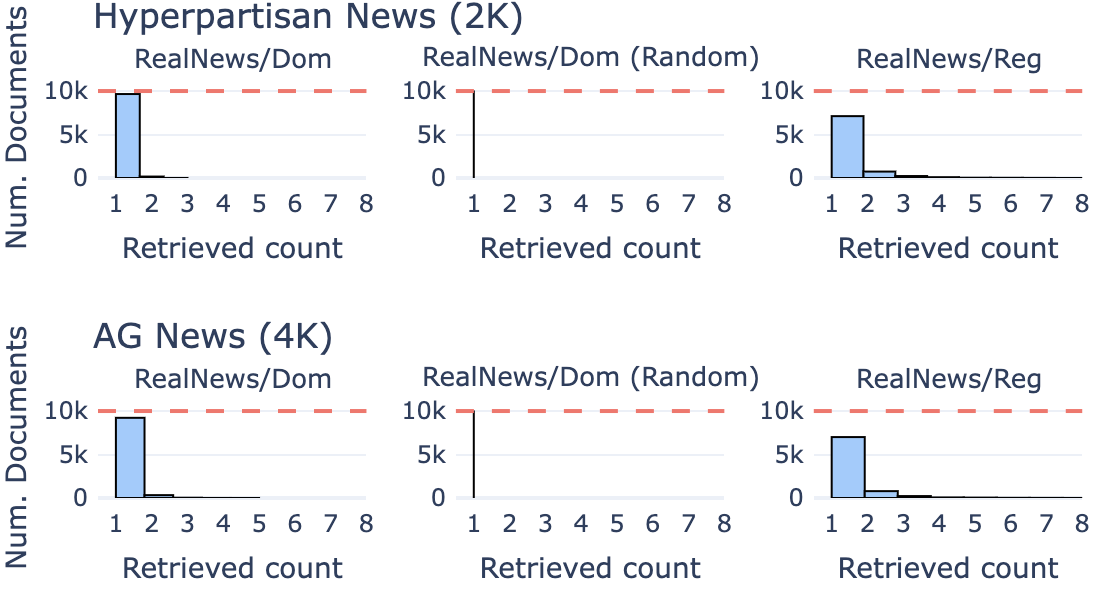}


\vspace{-1ex}
\centering
\caption{
Retrieval counts for \textsc{Hyperpartisan} and \textsc{AG News}. The red dashed line represents the theoretical max, where all retrieved documents are unique. Note that the Random histogram plot is always 1 hence shows up as a straight line. 
}
\vspace{-2ex}
\label{fig:retrieval-counts}
\end{figure}

\insection[]{RQ: is retrieval essential for content sourcing?}
We measure the importance of retrieval by selecting top-k documents randomly from the in-domain corpus \RealNewsDom. We observe in \autoref{tab:domain-shift} that retrieval using in-context learning queries plays a crucial role to the performance of \AGNews, as performance drops significantly in a random setting. \Hyperpartisan does not face such a drop. This matches our intuition in \autoref{tab:tasks} that task-inversion is the more challenging step for \Hyperpartisan, and a powerful LLM we can apply stylistic changes to most news articles. In both, Mauve suffers when entities no longer match \gold.

\insection[]{RQ: Do in-context queries retrieve redundant results?}
\autoref{fig:retrieval-counts} measures the overlap of top-50 retrieved documents from the 200 ICL queries, and finds that in most cases, the retrieved documents are unique, especially when using a large in-domain corpus. Thus, we can conclude that effective retrieval is important for the diversity of the synthetic dataset. 

\insection[]{RQ: Can \ours{} work effectively with relatively small corpora?}
In our main results \S\ref{sec:expts}, we assumed the existence of a large corpus, with minimum size of 0.9M documents. As noted, this corpus need not be unlabelled examples for our task; we were able to successfully generate customer reviews and product questions for \Humor, \Category{} and \Polarity{} tasks,  while retrieving from a corpus of product information (title and description). 

A potential problem with \ours{} is that corpuses of such massive size might be few in number. Thus, we compare the performance of \ours{} on \CMUMovies{} \cite{bamman-etal-2013-learning} which is between one to three orders of magnitude smaller than other corpora in \autoref{tab:student-deb}. In \autoref{tab:baselines-combined-detailed}, we see that \ours{} can perform suitably even with such relatively small corpora (42k movie plots). From the previous RQs, this suggests that the relevance of the corpus to the task is more important than the size of the corpus for the performance of \ours.


\section{Dense vs sparse retrieval in \ours}
\label{sec:bm25}

\begin{table}[!t]
\centering
\sisetup{round-mode=places, round-precision=1} 
\setlength{\tabcolsep}{2.5pt}
\footnotesize{

\begin{tabular}{
c  
C{0.7cm} 
C{0.7cm} 
C{0.7cm} 
C{0.7cm} 
C{0.7cm} 
C{0.7cm} 
C{0.7cm}  
}
\toprule
\textbf{Retriever}
& \AG          
& \Hyp        
& \ToI         
& \Cat     
& \Hum         
& \Pol 
& { }\multirow{2}{*}{\textbf{Avg.}}
\\ 
\textit{(Size)} 
& \text{($8$K)}    
& \text{($2$K)} 
& \text{($8$K)}  
& \text{($8$K)}   
& \text{($2$K)}  
& \text{($4$K)} 
&
\\ 
\midrule
\gold               &  91.0 &  93.2 &  82.5 &  81.5 &  93.1 &  95.3  & 89.43 \\
\toprule
\multicolumn{8}{c}{\underline{\textsc{\LLaMa{} Zero shot}}} \vspace{1ex} \\
\ContrieverShort    & \best{83.5} & 69.8 &  \best{74.5} &  \best{68.9} & \best{82.5} &  84.7 & \best{77.32} \\
\BM                 & 83.2 & \best{74.2} & 70.7   & 57.6 & 78.5   & \best{85.4}  & 74.93  \\
[1.0ex]
\toprule
\multicolumn{8}{c}{\underline{\textsc{\Claude{} Zero shot}}} \vspace{1ex} \\
\ContrieverShort    & \best{83.9} &  \best{72.3} &  \best{71.8}  &  \best{66.8} &  62.1 &  88.7  & \best{74.29} \\
\BM                 & 83.2 & 57.2   &  69.8  & 53.7 & \best{73.9}   & \best{91.8}  & 71.60  \\
[1.0ex]
\toprule
\multicolumn{8}{c}{\underline{\textsc{\LLaMa{} 3-shot \RetrICL}}} \vspace{1ex} \\
\ContrieverShort    & \best{83.0} &  \best{78.5} &  \best{73.3} & \best{72.4} & \best{90.2}  &  \best{91.0}   & \best{81.38} \\
\BM                 & 82.1 & 77.9 & 71.9   & 65.4   & 87.5   & 87.4  & 78.69  \\
[1.0ex]
\toprule
\multicolumn{8}{c}{\underline{\textsc{\Claude{} 3-shot \RetrICL}}} \vspace{1ex} \\
\ContrieverShort    & \best{83.7}  &  72.3 &  \best{72.8} &  \best{65.4} &  \best{83.4} & \best{91.3}  & \best{78.16} \\
\BM                 & 83.0 & \best{73.5}   & 70.0   & 52.4 & 82.4 & 90.7  & 75.34  \\
\bottomrule
\end{tabular}
}
\vspace{-1ex}
\caption{
Test accuracy after distilling a \DeBERTa{} student on a dataset generated by \ours{}. Retrieval is done using \BM{} and \Contriever{}. We use the same seed set and corpus as \autoref{tab:corpus}. We train 5 student models and report the mean accuracy, rerunning all 5 in case of std $\ge$ 6.0. The top two subsections consider zero-shot synthesis and bottom two considers 3-shot \RetrICL{} variant. We \textbf{bold} the best result in each subsection. \Contriever{} numbers are reproduced from \autoref{tab:student-deb}.
}
\vspace{-3ex}
\label{tab:bm25-student}
\end{table}

So far, a single dense retriever (\Contriever) has been used for the content sourcing step by using a bi-encoder approach \citep{lee-etal-2019-latent, chen-etal-2017-reading}. We embed both the input in-context covariate and each corpus document, and then rank results based on cosine similarity. In \S\ref{sec:expts}, we retrieved $k=500$ documents for each in-context example and after filtering, randomly sampled among these to produce a grounded set of documents on which we apply our task inversion strategy \RetrICL. 

In this section we explore how changing the retrieval model affects the content sourcing stage and its downstream effects. Keeping other parts of the process the same, we switch \Contriever{} to \BM{} Okapi \citep{bm25}, a popular \textit{sparse} retrieval method. Dense retrievers like \Contriever{} perform a semantic match between the query and document, whereas \BM{} performs only a lexical match based on inverse term frequencies, with no understanding of semantics. Additionally, \BM{} outputs a score which is an unbounded positive number, thus we are unable to use meaningful thresholds to bound the similarity in our \RetrICL{} approach. Instead, we construct the \RetrICL{} in-context set using the top-2 retrieved contexts for each ICL example and without applying the filter.

We expect that picking semantically similar information is more important to \ours{} since we include a task inversion step, which intends to change the tone and lexical structure of the text while preserving its semantics. Thus, we want contexts which are semantically related to \gold{} data, to which we can apply stylistic or formatting transformations using a task-inversion prompt to bring it closer to \gold. 

Surprisingly, in \autoref{tab:baselines-combined-detailed} we see that while intrinsic diversity from \BM-retrieved documents is often worse than \Contriever, they both generate equally human-like text. However, comparing the \DeBERTa{} accuracy of \Contriever{} and \BM in \autoref{tab:bm25-student}, we see that a strong student model trained on a dataset obtained from the dense-retrieved document set consistently outperforms the sparse retriever \BM, which might be due to the filtering step we introduce in \RetrICL. This filtering step leads to a reduction in mislabelling stemming from retrieving contexts that belong do a different class. Due to this, we conclude that dense retrieval models are potentially more suitable for \ours.
\section{Varying number of in-context examples in \RetrICL}
\label{sec:num-icl}
\begin{figure*}[!ht]
\includegraphics[width=0.31\textwidth]{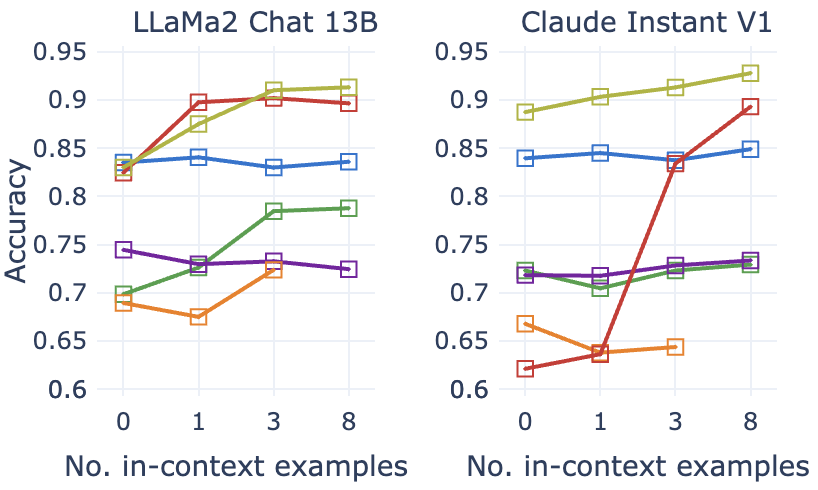} \hfill \includegraphics[width=0.31\textwidth]{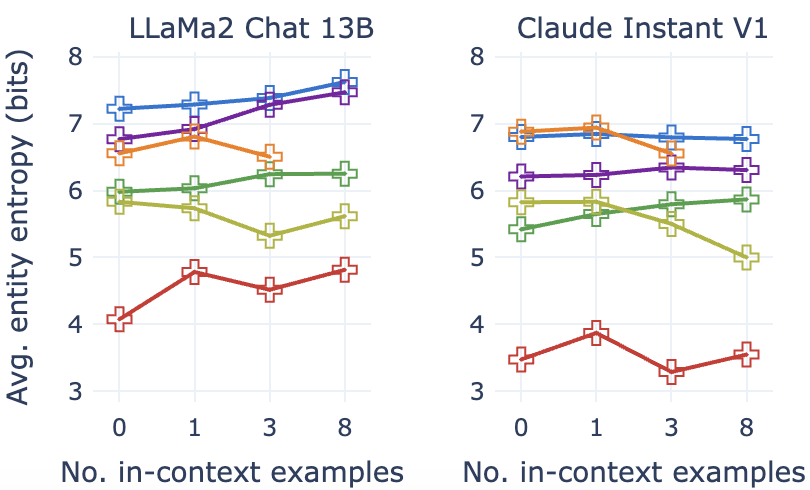} \hfill \includegraphics[width=0.31\textwidth]{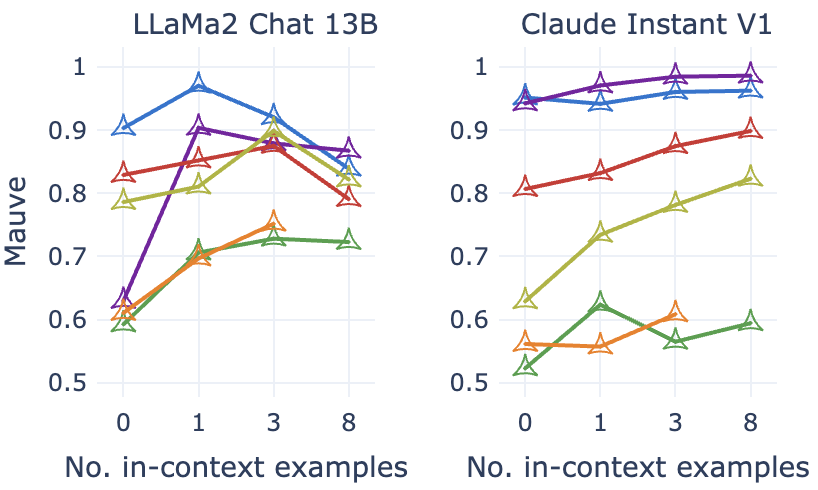}
\vspace{-0.5ex}
\centering
\caption{
Left: \DeBERTa{} test accuracy \higherbetter{}, center: entity entropy \higherbetter{}, right: Mauve \higherbetter{} for \ours{} \RetrICL. We vary the number of in-context examples from 0 to 8. Teacher LLMs \LLaMaFull{} and \ClaudeFull{} are compared on 6 tasks: 
\textcolor[HTML]{1976d2}{{\AGNews}}, 
\textcolor[HTML]{43a047}{\Hyperpartisan}, 
\textcolor[HTML]{7b1fa2}{\ToIHeadlines}, 
\textcolor[HTML]{f57c00}{\Category}, 
\textcolor[HTML]{d32f2f}{\Humor} and 
\textcolor[HTML]{827717}{\Polarity}.
We do not report \textcolor[HTML]{f57c00}{\Category} 8-shot due to model failures.
}
\vspace{-2ex}
\label{fig:num_icl-other-mets}
\end{figure*}
\begin{figure*}[!t]
\hfill \includegraphics[width=0.45\textwidth]{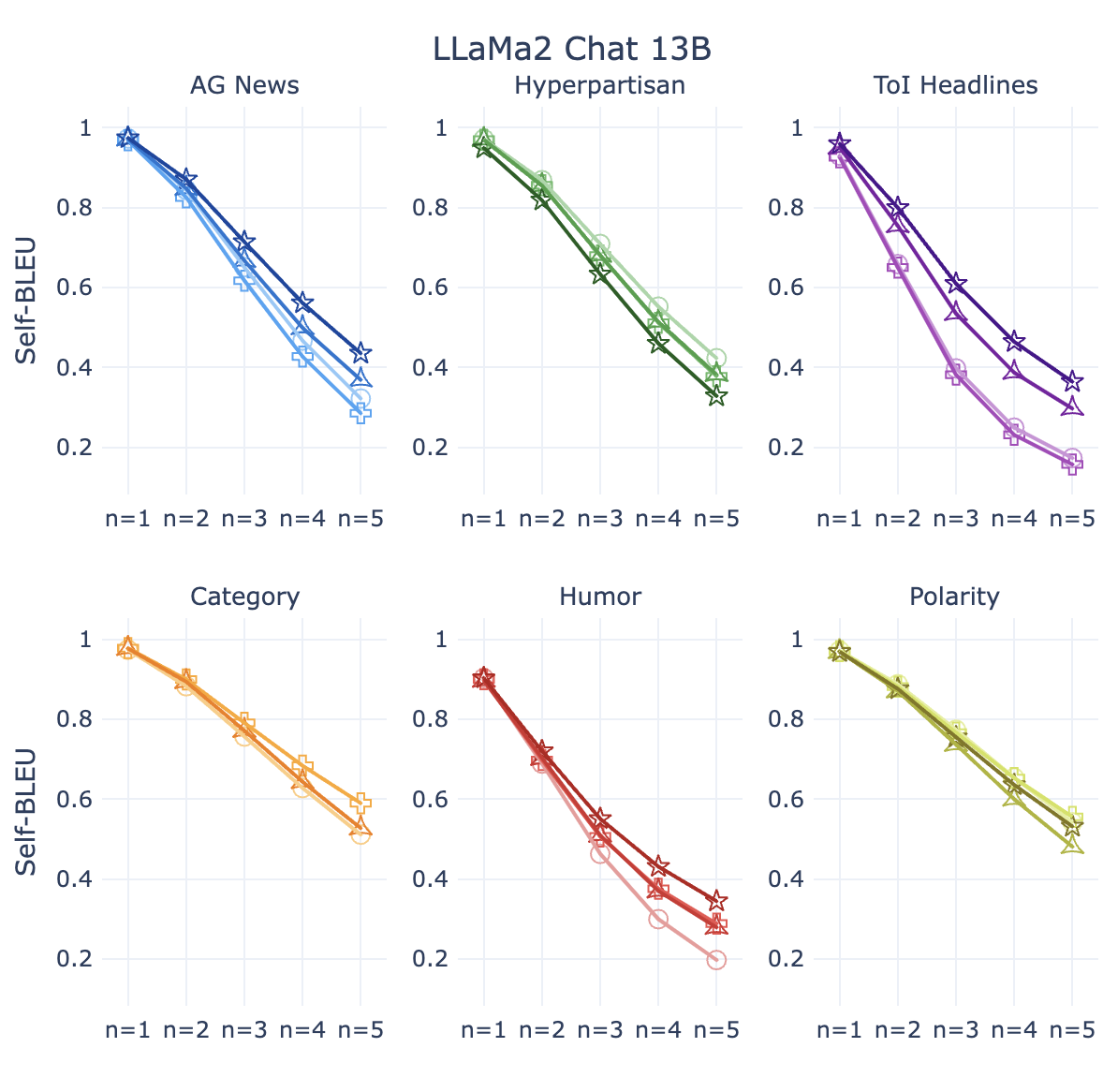} \hfill \includegraphics[width=0.45\textwidth]{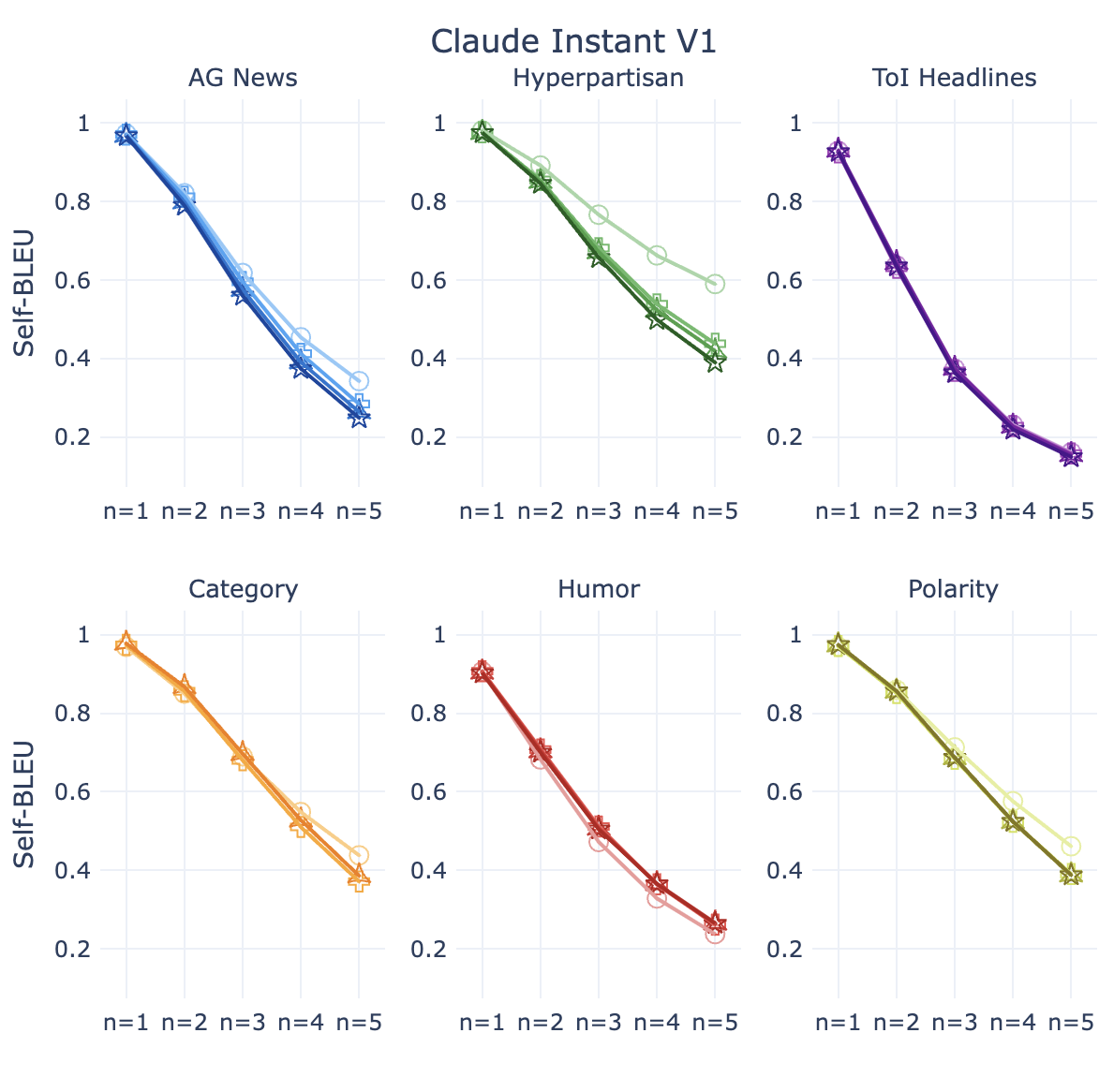} \hfill 
\vspace{-0.5ex}
\centering
\caption{
Lexical diversity i.e. Self-BLEU \lowerbetter{} ngrams n=1-5, when varying the number of in-context examples for \ours{} \RetrICL. We compare of teacher LLMs \LLaMaFull{} (left) and \ClaudeFull{} (right). Notation: 0-shot ($\bullet$), 1-shot (+), 3-shot ($\vartriangle$), 8-shot ($\bigstar$). Darker shade implies more in-context examples.
}
\vspace{-0.5ex}
\label{fig:num_icl-self_bleu}
\end{figure*}

The use of in-context examples in the \RetrICL{} variant of \ours{} leads to significant improvements in intrinsic and distillation metrics, as we saw in \S\ref{sec:expts}. Here, we do a deeper analysis on whether continually increasing the number of in-context examples yields a positive benefit.

In \autoref{fig:num_icl-other-mets} we look at the \DeBERTa{} accuracy, entity entropy and MAUVE for our datasets with different numbers of in-context learning examples. We see that adding even a single in-context example can greatly increase the performance of all three metrics. However, no particular number of in-context examples consistently outperforms. For \Claude{}, adding more in-context examples (up to 8) seems to always provide benefit, whereas with \LLaMa, we observe a peak and then reduction. Thus, the optimal number of in-context learning examples is a task dependent hyperparameter. 

\autoref{fig:num_icl-self_bleu} shows the lexical diversity i.e. Self-BLEU across datasets and number of in-context examples. As in \S\ref{sec:expts} we observed that using in-context examples is neither positively nor negatively correlated with a lower Self-BLEU, despite using nucleus sampling with $p=0.9$. This may be because for all number of shots, task inversion is performed from a single source context and thus the generation does not divert significantly from the unique n-grams of the context. Thus we conclude that to affect lexical diversity, the number of in-context learning examples has no effect and we must instead focus on changing the retrieved contexts, perhaps by using a different retrieval model.

\section{Task inversion prompts and label verbalizations}
\label{sec:prompts}

Here we discuss the prompt templates and verbalizations that we use for the task inversion step for both \fewgen{} and \ours. We use descriptive verbalizations as compared to the target label. 

Additionally in the prompt, we place the retrieved document near the end, as prior work indicates that intermediate placements degrade LLM recall \citep{liu-etal:2023:arxiv}. 

LLMs have a fixed window-size for conditional generation, so excessively long documents are truncated (from the end) up to $r_{max}=500$ tokens. This reserves the remaining window for in-context learning.

\subsection{\Hyperpartisan}

\Hyperpartisan{} is the task of detecting political bias in a news article. 
In transforming the retrieved news article \paramnorm{article_retr[k]} to one with such bias, typically there is the addition of mocking commentary and harsh political language which deeply criticizes the subject such as a person, policy or political event. On the other hand, articles in the opposite class gives a well-rounded opinion with a neutral tone. We include a length-attribute to ensure a long generation of one or two paragraphs. 
\begin{table}[!h]
\centering
\tiny{
\setlength{\tabcolsep}{3pt}
\begin{tabular}{C{1cm}p{0.4\textwidth}}
\toprule
\bf{Label} & \cellhalign{\bf{Verbalization}} \\
\midrule
true & harsh political language, using a mocking tone and toxic commentary \\ 
\midrule
false & neutral language, using a reasonable tone and politically correct commentary \\
\bottomrule
\end{tabular}
\vspace{-1ex}
}
\caption{
Task-inversion verbalizations for \Hyperpartisan.
}
\vspace{-3ex}
\label{tab:examples}
\end{table}

\begin{prompt}[title={Prompt \thetcbcounter: \Hyperpartisan{} \fewgen}, label=prompt:dataset]
\promptsubsection{In-context example} \\ \prompttext{
Write a single news article using \param{{label}}. The written article should be 2 to 3 paragraphs long.
\\ News Article: \param{{icl[gold_text]}}
}
\\ \promptsubsection{Prompt} \\ \prompttext{
Write a single news article using \param{{label}}. The written article should be 2 to 3 paragraphs long.
\\ News Article:
}
\end{prompt}
\begin{prompt}[title={Prompt \thetcbcounter: \Hyperpartisan{} \ours{} \RetrICL}, label=prompt:dataset]
\promptsubsection{In-context example} \\ \prompttext{
News Article:
\param{{icl[article_retr]}}
\\ Rewrite the above news article using \param{{label}}. The rewritten article should be 2 to 3 paragraphs long. 
\\ Rewritten Article: \param{{icl[gold_text]}}
}
\\ \promptsubsection{Prompt} \\ \prompttext{
News Article: 
\param{{article_retr[k]}}
\\ Rewrite the above news article using \param{{label}}. The rewritten article should be 2 to 3 paragraphs long. 
\\ Rewritten Article: 
}
\end{prompt}
\begin{prompt}[title={Prompt \thetcbcounter: \Hyperpartisan{} \ours{} \NoRetrICL}, label=prompt:dataset]
\promptsubsection{In-context example} \\ \prompttext{
Rewritten Article: 
\param{{icl[gold_text]}}
}
\\ \promptsubsection{Prompt} \\ \prompttext{
News Article:
\param{{article_retr[k]}}
\\ Rewrite the above news article using \param{{label}}. The rewritten article should be 2 to 3 paragraphs long. 
\\ Rewritten Article: 
}
\end{prompt}





\subsection{\ToIHeadlines}
\ToIHeadlines{} is a topic classification dataset of regional news headlines in India. Here we attempt to refine the retrieved news article by summarizing it into a short headline. We use verbalizations of the content of each class, as example generation here involves summarizing the content. We add an ``India'' location-attribute to guide the LLM generations to include regionalization to the Indian subcontinent. A length-attribute is included to restrict the length to one sentence.
\begin{table}[!h]
\centering
\tiny{
\setlength{\tabcolsep}{3pt}
\begin{tabular}{C{2.5cm}p{0.3\textwidth}}
\toprule
\bf{Label} & \cellhalign{\bf{Verbalization}} \\
\midrule
sports & sports in India \\ 
\midrule
life-style &  health and lifestyle trends in India \\ 
\midrule
education & Indian examinations and education \\ 
\midrule
entertainment & the Indian entertainment industry \\ 
\midrule
business & business-related developments in India \\ 
\midrule
city & ongoing matters in any Indian city \\ 
\midrule
environment & environment-related events in Indian cities \\ 
\midrule
tech & technology news and the tech industry in India \\ 
\midrule
elections & elections and politics in India \\ 
\midrule
world & international news and events outside of India \\ 
\bottomrule
\end{tabular}
\vspace{-1ex}
}
\caption{
Task-inversion verbalizations for \ToIHeadlines.
}
\vspace{-3ex}
\label{tab:examples}
\end{table}

\begin{prompt}[title={Prompt \thetcbcounter: \ToIHeadlines{} \fewgen}, label=prompt:toi]
\promptsubsection{In-context example} \\ \prompttext{
Write a headline for a news article about \param{{label}}. The headline should be a single sentence.
\\ Headline: \param{{icl[gold_text]}}
}
\\ \promptsubsection{Prompt} \\ \prompttext{
Write a headline for a news article about \param{{label}}. The headline should be a single sentence.
\\ Headline: 
}
\end{prompt}
\begin{prompt}[title={Prompt \thetcbcounter: \ToIHeadlines{} \ours{} \RetrICL}, label=prompt:dataset]
\promptsubsection{In-context example} \\ \prompttext{
News Article:
\param{{icl[article_retr]}}
\\ Write a headline for the above news article about \param{{label}}. The headline should be a single sentence.
\\ Headline: \param{{icl[gold_text]}}
}
\\ \promptsubsection{Prompt} \\ \prompttext{
News Article:
\param{{article_retr[k]}}
\\ Write a headline for the above news article about \param{{label}}. The headline should be a single sentence. 
\\ Headline:
}
\end{prompt}
\begin{prompt}[title={Prompt \thetcbcounter: \ToIHeadlines{} \ours{} \NoRetrICL}, label=prompt:dataset]
\promptsubsection{In-context example} \\ \prompttext{
Headline: \param{{icl[article_retr]}}
}
\\ \promptsubsection{Prompt} \\ \prompttext{
News Article:
\param{{article_retr[k]}}
\\ Write a headline for the above news article about \param{{label}}. The headline should be a single sentence. 
\\ Headline: 
}
\end{prompt}

\subsection{\AGNews}

We consider task inversion for the \AGNews{} dataset to be generation of news summaries. We do not specify location modifiers as most \gold{} summaries are from US news. We add a length-attribute to restrict the output one or two sentences.

\begin{table}[!h]
\centering
\tiny{
\setlength{\tabcolsep}{3pt}
\begin{tabular}{C{1cm}p{0.4\textwidth}}
\toprule
\bf{Label} & \cellhalign{\bf{Verbalization}} \\
\midrule
Business & companies, industries, markets, trade, investments, entrepreneurship, economic policies, and other business-related developments \\ 
\midrule
World & international news, such as politics, diplomacy, conflicts, global events, international relations, human rights issues, and significant global trends \\ 
\midrule
Sci/Tech & scientific discoveries, technological advancements, innovations, research breakthroughs \\ 
\midrule
Sports & professional sports leagues, major tournaments, athletes, teams, match results, player transfers, coaching changes, sports-related controversies \\ 
\bottomrule
\end{tabular}
\vspace{-1ex}
}
\caption{
Task-inversion verbalizations for \AGNews.
}
\vspace{-3ex}
\label{tab:examples}
\end{table}

\begin{prompt}[title={Prompt \thetcbcounter: \AGNews{} \fewgen}, label=prompt:agnews]
\promptsubsection{In-context example} \\ \prompttext{
Write a summary for a news article about \param{{label}}. The summary should be one or two short sentences.
\\ Summary: \param{{icl[gold_text]}}
}
\\ \promptsubsection{Prompt} \\ \prompttext{
Write a summary for a news article about \param{{label}}. The summary should be one or two short sentences.
\\Summary:
}
\end{prompt}
\begin{prompt}[title={Prompt \thetcbcounter: \AGNews{} \ours{} \RetrICL}, label=prompt:dataset]
\promptsubsection{In-context example} \\ \prompttext{News Article:
\param{{icl[article_retr]}}
\\Write a summary for the above news article about \param{{label}}. The summary should be one or two short sentences.
\\Summary: \param{{icl[gold_text]}}
}
\\ \promptsubsection{Prompt} \\ \prompttext{
News Article:
\param{{article_retr[k]}}
\\Write a summary for the above news article about \param{{label}}. The summary should be one or two short sentences. 
\\Summary:
}
\end{prompt}
\begin{prompt}[title={Prompt \thetcbcounter: \AGNews{} \ours{} \NoRetrICL}, label=prompt:dataset]
\promptsubsection{In-context example} \\ \prompttext{
Summary: \param{{icl[gold_text]}}
}
\\ \promptsubsection{Prompt} \\ \prompttext{
News Article:
\param{{article_retr[k]}}
\\Write a summary for the above news article about \param{{label}}. The summary should be one or two short sentences. 
\\Summary:
}
\end{prompt}

\subsection{\Category}

In the \Category{} dataset, we determine the product category from a review written by a user about products on a major e-commerce website. For task inversion in \ours{} we must retrieve a product and prompt the frozen LLM to generate a user review within the same product-category as the retrieval query. Thus, we include a style-attribute to allow minor typos in the generation and restrict to a few sentences using a length-attribute.

\begin{table}[!h]
\centering
\tiny{
\setlength{\tabcolsep}{3pt}
\begin{tabular}{C{2.5cm}p{0.3\textwidth}}
\toprule
\bf{Label} & \cellhalign{\bf{Verbalization}} \\
\midrule

\detokenize{magazines} & magazines or periodicals covering various topics \\ 
\midrule
\detokenize{camera_photo} & photography gear including cameras, lenses, accessories, or photo editing tools \\ 
\midrule
\detokenize{office_products} & office supplies or equipment for professional and home office setups \\ 
\midrule
\detokenize{kitchen} & kitchenware, appliances, or culinary tools for cooking and dining \\ 
\midrule
\detokenize{cell_phones_service} & cell phone service accessories or service plans for communication and connectivity \\ 
\midrule
\detokenize{computer_video_games} & computers, gaming consoles, video games, or related accessories \\ 
\midrule
\detokenize{grocery_and_gourmet_food} & groceries, fruits and vegetables, gourmet treats, or specialty food items \\ 
\midrule
\detokenize{tools_hardware} & tools, hardware, or equipment for DIY projects and home repairs \\ 
\midrule
\detokenize{automotive} & auto parts, accessories, or tools for vehicle maintenance and enhancements \\ 
\midrule
\detokenize{music_album} & music albums spanning various genres and artists \\ 
\midrule
\detokenize{health_and_personal_care} & healthcare products, personal care items, or wellness essentials \\ 
\midrule
\detokenize{electronics} & electronic devices, gadgets, personal tech, or home electronics \\ 
\midrule
\detokenize{outdoor_living} & products for outdoor activities, gardening, or patio living \\ 
\midrule
\detokenize{video} & movies, TV shows, and documentaries spanning various genres and artists \\ 
\midrule
\detokenize{apparel} & clothing including casual wear, formal attire, seasonal outfits, activewear, or fashion accessories for men, women, and children \\ 
\midrule
\detokenize{toys_games} & fun or educational toys and games for kids of all ages \\ 
\midrule
\detokenize{sports_outdoors} & products for various sports and outdoor activities \\ 
\midrule
\detokenize{books} & books in various genres and formats \\ 
\midrule
\detokenize{software} & computer software for productivity or gaming covering either personal or professional needs \\ 
\midrule
\detokenize{baby} & baby essentials, gear, or toys for infants and toddlers \\ 
\midrule
\detokenize{musical_and_instruments} & musical instruments, accessories, or music production equipment \\ 
\midrule
\detokenize{beauty} & beauty products, cosmetics, or skincare essentials, makeup, hair care, fragrances, or grooming essentials \\ 
\midrule
\detokenize{jewelry_and_watches} & watches or jewelry pieces such as necklaces, bracelets, earrings, or rings, crafted in precious metals or adorned with gemstones for special occasions \\ 

\bottomrule
\end{tabular}
\vspace{-1ex}
}
\caption{
Task-inversion verbalizations for \Category.
}
\vspace{-3ex}
\label{tab:examples}
\end{table}

\begin{prompt}[title={Prompt \thetcbcounter: \Category{} \fewgen}, label=prompt:dataset]
\promptsubsection{In-context example} \\ \prompttext{
Write a product review about a product which is in the category of \param{{label}}. Include relevant product details. The review should only be a single short sentence, or a single paragraph of 3 to 4 sentences. Add very minor typos.
\\ Review: \param{{icl[gold_text]}}
}
\\ \promptsubsection{Prompt} \\ \prompttext{
Write a product review about a product which is in the category of \param{{label}}. Include relevant product details. The review should only be a single short sentence, or a single paragraph of 3 to 4 sentences. Add very minor typos.
\\ Review: 
}
\end{prompt}
\begin{prompt}[title={Prompt \thetcbcounter: \Category{} \ours{} \RetrICL}, label=prompt:dataset]
\promptsubsection{In-context example} \\ \prompttext{
Product details:
\param{{icl[product_retr]}}
\\ Write a product review about the above product which is in the category of \param{{label}}. Include relevant product details which are mentioned above. The review should only be a single short sentence, or a single paragraph of 3 to 4 sentences. Add very minor typos.
\\ Review: \param{{icl[gold_text]}}
}
\\ \promptsubsection{Prompt} \\ \prompttext{
Product details:
\param{{product_retr[k]}}
\\ Write a product review about the above product which is in the category of \param{{label}}. Include relevant product details which are mentioned above. The review should only be a single short sentence, or a single paragraph of 3 to 4 sentences. Add very minor typos. 
\\ Review:
}
\end{prompt}
\begin{prompt}[title={Prompt \thetcbcounter: \Category{} \ours{} \NoRetrICL}, label=prompt:dataset]
\promptsubsection{In-context example} \\ \prompttext{
Review: \param{{icl[gold_text]}}
}
\\ \promptsubsection{Prompt} \\ \prompttext{
Product details:
\label{{product_retr[k]}}
\\ Write a product review about the above product which is in the category of \param{{label}}. Include relevant product details which are mentioned above. The review should only be a single short sentence, or a single paragraph of 3 to 4 sentences. Add very minor typos. 
\\ Review:
}
\end{prompt}




\subsection{\Humor}
Asking humorous product questions is a challenge of the LLM's task inversion capabilities, as it must generate a question which is funny from the retrieved product. Not all products have obvious humorous characteristics, thus the generation requires some ingenuity. We restrict the output to only the question to prevent explanations or extraneous product generations from the LLM.

\begin{table}[!h]
\centering
\tiny{
\setlength{\tabcolsep}{3pt}
\begin{tabular}{C{2.5cm}C{0.3\textwidth}}
\toprule
\bf{Label} & \cellhalign{\bf{Verbalization}} \\
\midrule
\detokenize{humorous} & humorous \\ 
\midrule
\detokenize{non_humorous} & solemn \\ 
\bottomrule
\end{tabular}
\vspace{-1ex}
}
\caption{
Task inversion verbalizations for \Humor.
}
\vspace{-3ex}
\label{tab:examples}
\end{table}

\begin{prompt}[title={Prompt \thetcbcounter: \Humor{} \fewgen}, label=prompt:dataset]
\promptsubsection{In-context example} \\ \prompttext{
Write a short \param{{label}} question about a product. Only include the question.
\\ Product Question: \param{{icl[gold_text]}}
}
\\ \promptsubsection{Prompt} \\ \prompttext{
Write a short \param{{label}} question about a product. Only include the question.
\\ Product Question: 
}
\end{prompt}
\begin{prompt}[title={Prompt \thetcbcounter: \Humor{} \ours{} \RetrICL}, label=prompt:dataset]
\promptsubsection{In-context example} \\ \prompttext{
Product details:
\param{{icl[product_retr]}}
\\ Write a short \param{{label}} question about the above product. Only include the question.
\\ Product Question: \param{{icl[gold_text]}}
}
\\ \promptsubsection{Prompt} \\ \prompttext{
Product details:
\param{{product_retr[k]}}
\\ Write a short \param{{label}} question about the above product. Only include the question. 
\\ Product Question:
}
\end{prompt}
\begin{prompt}[title={Prompt \thetcbcounter: \Humor{} \ours{} \NoRetrICL}, label=prompt:dataset]
\promptsubsection{In-context example} \\ \prompttext{
Product Question: \param{{icl[gold_text]}}
}
\\ \promptsubsection{Prompt} \\ \prompttext{
Product details:
\param{{product_retr[k]}}
\\ Write a short \param{{label}} question about the above product. Only include the question. 
\\ Product Question:
}
\end{prompt}

\subsection{\Polarity}
\Polarity{} is a sentiment classification task for reviews of products on a major e-commerce website. In \ours, the difficulty is increased as we must generate a review from a product. For task inversion, we prompt the LLM to generate a review which can have either positive or negative sentiment and include details from the retrieved product. As with \Category, we allow typos and restrict the length to a few sentences using a length-attribute in the prompt.

\begin{table}[!h]
\centering
\tiny{
\setlength{\tabcolsep}{3pt}
\begin{tabular}{C{1cm}p{0.4\textwidth}}
\toprule
\bf{Label} & \cellhalign{\bf{Verbalization}} \\
\midrule
\detokenize{positive} & what the reviewer liked about the product, how the reviewer found it easy to use the product, or the reviewer's positive experience with the product \\ 
\midrule
\detokenize{negative} & what the reviewer disliked about the product, how the reviewer found it challenging to use the product, or the reviewer's negative experience with the product \\ 
\bottomrule
\end{tabular}
\vspace{-1ex}
}
\caption{
Task inversion verbalizations for \Polarity.
}
\vspace{-3ex}
\label{tab:examples}
\end{table}

\begin{prompt}[title={Prompt \thetcbcounter: \Polarity{} \fewgen}, label=prompt:dataset]
\promptsubsection{In-context example} \\ \prompttext{
Write a review about a product which discusses \param{{label}}. Include relevant product details. The review should only be a single short sentence, or a single paragraph of 3 to 4 sentences. Add very minor typos.
\\ Review: \param{{icl[gold_text]}}
}
\\ \promptsubsection{Prompt} \\ \prompttext{
Write a review about a product which discusses \param{{label}}. Include relevant product details. The review should only be a single short sentence, or a single paragraph of 3 to 4 sentences. Add very minor typos.
\\ Review: 
}
\end{prompt}
\begin{prompt}[title={Prompt \thetcbcounter: \Polarity{} \ours{} \RetrICL}, label=prompt:dataset]
\promptsubsection{In-context example} \\ \prompttext{
Product details:
\param{{icl[product_retr]}}
\\ Write a review about the above product which discusses \param{{label}}. Include relevant product details which are mentioned above. The review should only be a single short sentence, or a single paragraph of 3 to 4 sentences. Add very minor typos.
\\ Review: \param{{icl[gold_text]}}
}
\\ \promptsubsection{Prompt} \\ \prompttext{
Product details:
\param{{product_retr[k]}}
\\ Write a review about the above product which discusses \param{{label}}. Include relevant product details which are mentioned above. The review should only be a single short sentence, or a single paragraph of 3 to 4 sentences. Add very minor typos. 
\\ Review: 
}
\end{prompt}
\begin{prompt}[title={Prompt \thetcbcounter: \Polarity{} \ours{} \NoRetrICL}, label=prompt:dataset]
\promptsubsection{In-context example} \\ \prompttext{
Review: \param{{icl[gold_text]}}
}
\\ \promptsubsection{Prompt} \\ \prompttext{
Product details:
\param{{product_retr[k]}}
\\ Write a review about the above product which discusses \param{{label}}. Include relevant product details which are mentioned above. The review should only be a single short sentence, or a single paragraph of 3 to 4 sentences. Add very minor typos. 
\\ Review:
}
\end{prompt}

%
\subsection{\IMDb}
\IMDb{} is a review-sentiment classification task. As with other review tasks, in the task inversion step we prompt the LLM to generate a review in either positive or negative sentiment. The context used by \ours{} is the plotline of a movie from \CMUMovies. As with \Category{} and \Polarity, we allow typos and restrict the length to a few sentences using a length-attribute in the prompt.

\begin{table}[!h]
\centering
\tiny{
\setlength{\tabcolsep}{3pt}
\begin{tabular}{C{1cm}p{0.4\textwidth}}
\toprule
\bf{Label} & \cellhalign{\bf{Verbalization}} \\
\midrule
\detokenize{positive} & what the reviewer liked about the movie \\ 
\midrule
\detokenize{negative} & what the reviewer disliked about the movie \\ 
\bottomrule
\end{tabular}
\vspace{-1ex}
}
\caption{
Task inversion verbalizations for \IMDb.
}
\vspace{-3ex}
\label{tab:examples}
\end{table}

\begin{prompt}[title={Prompt \thetcbcounter: \IMDb{} \fewgen}, label=prompt:dataset]
\promptsubsection{In-context example} \\ \prompttext{
Write a review which discusses \param{{label}}. Include relevant details about the movie. The review should only be a single short sentence, or a single paragraph of 3 to 4 sentences. Add very minor typos.
\\ Review: \param{{icl[gold_text]}}
}
\\ \promptsubsection{Prompt} \\ \prompttext{
Write a review which discusses \param{{label}}. Include relevant details about the movie. The review should only be a single short sentence, or a single paragraph of 3 to 4 sentences. Add very minor typos.
\\ Review: 
}
\end{prompt}
\begin{prompt}[title={Prompt \thetcbcounter: \IMDb{} \ours{} \RetrICL}, label=prompt:dataset]
\promptsubsection{In-context example} \\ \prompttext{
Movie details:
\param{{icl[plotline_retr]}}
\\ Write a review which discusses \param{{label}}. Include relevant details about the movie which are mentioned above. The review should only be a single short sentence, or a single paragraph of 3 to 4 sentences. Add very minor typos.
\\ Review: \param{{icl[gold_text]}}
}
\\ \promptsubsection{Prompt} \\ \prompttext{
Movie details:
\param{{plotline_retr[k]}}
\\ Write a review which discusses \param{{label}}. Include relevant details about the movie which are mentioned above. The review should only be a single short sentence, or a single paragraph of 3 to 4 sentences. Add very minor typos.
\\ Review: 
}
\end{prompt}
\begin{prompt}[title={Prompt \thetcbcounter: \IMDb{} \ours{} \NoRetrICL}, label=prompt:dataset]
\promptsubsection{In-context example} \\ \prompttext{
Review: \param{{icl[gold_text]}}
}
\\ \promptsubsection{Prompt} \\ \prompttext{
Movie details:
\param{{plotline_retr[k]}}
\\ Write a review which discusses \param{{label}}. Include relevant details about the movie which are mentioned above. The review should only be a single short sentence, or a single paragraph of 3 to 4 sentences. Add very minor typos. 
\\ Review:
}
\end{prompt}


\subsection{\SST}
\SST{} is another review-sentiment classification task, however the examples are partial sentences from movie reviews which were extracted such that they contain the sentiment-heavy phrases. This, during the task inversion we prompt the Teacher LLM to generate a partial review sentence in either positive or negative sentiment. The context used by \ours{} is the plotline of a movie from \CMUMovies. We allow typos and restrict the length to one sentence using a length-attribute in the prompt.

\begin{table}[!h]
\centering
\tiny{
\setlength{\tabcolsep}{3pt}
\begin{tabular}{C{1cm}p{0.4\textwidth}}
\toprule
\bf{Label} & \cellhalign{\bf{Verbalization}} \\
\midrule
\detokenize{positive} & what the reviewer liked about the movie \\ 
\midrule
\detokenize{negative} & what the reviewer disliked about the movie \\ 
\bottomrule
\end{tabular}
\vspace{-1ex}
}
\caption{
Task inversion verbalizations for \SST.
}
\vspace{-3ex}
\label{tab:examples}
\end{table}

\begin{prompt}[title={Prompt \thetcbcounter: \SST{} \fewgen}, label=prompt:dataset]
\promptsubsection{In-context example} \\ \prompttext{
Write a single sentence from a review which discusses \param{{label}}. Include relevant details about the movie. The review should only be a single short sentence. Add very minor typos.
\\ Review: \param{{icl[gold_text]}}
}
\\ \promptsubsection{Prompt} \\ \prompttext{
Write a single sentence from a review which discusses \param{{label}}. Include relevant details about the movie. The review should only be a single short sentence. Add very minor typos.
\\ Review: 
}
\end{prompt}
\begin{prompt}[title={Prompt \thetcbcounter: \SST{} \ours{} \RetrICL}, label=prompt:dataset]
\promptsubsection{In-context example} \\ \prompttext{
Movie details:
\param{{icl[plotline_retr]}}
\\ Write a single sentence from a review which discusses \param{{label}}. Include relevant details about the movie which are mentioned above. The review should only be a single short sentence. Add very minor typos.
\\ Review: \param{{icl[gold_text]}}
}
\\ \promptsubsection{Prompt} \\ \prompttext{
Movie details:
\param{{plotline_retr[k]}}
\\ Write a single sentence from a review which discusses \param{{label}}. Include relevant details about the movie which are mentioned above. The review should only be a single short sentence. Add very minor typos.
\\ Review: 
}
\end{prompt}
\begin{prompt}[title={Prompt \thetcbcounter: \SST{} \ours{} \NoRetrICL}, label=prompt:dataset]
\promptsubsection{In-context example} \\ \prompttext{
Review: \param{{icl[gold_text]}}
}
\\ \promptsubsection{Prompt} \\ \prompttext{
Movie details:
\param{{plotline_retr[k]}}
\\ Write a single sentence from a review which discusses \param{{label}}. Include relevant details about the movie which are mentioned above. The review should only be a single short sentence. Add very minor typos. 
\\ Review:
}
\end{prompt}
\section{Example generations}
\label{sec:examples}

Here we showcase examples from the best-performing \ours{} approach (3-shot \NoRetrICL{} using \LLaMaFull{}) for each of our 6 tasks. For brevity, we do not show the ICL examples, only the retrieved article and generated text. 


\genex{\Hyperpartisan}
{hyperpartisan}{
News Article:

Gov. Andrew Cuomo Wednesdy afternoon staged a press conference to blast federal immigration officials, threatening to sue if agents don't stop what he described as "reckless and unconstitutional enforcement" practices.

Immigration and Customs Enforcement Deputy Director Thomas D. Homan responded hours later with a six-paragraph statement that accused Cuomo of "grandstanding."

Cuomo cited a raid last week on a farm in Rome, N.Y. Immigration and Customs Enforcement agents entered John Collins' property and detained a worker, Marcial DeLeon Aguilar, without presenting a warrant to Collins. That worker was later arrested and is still being detained.

"The reckless and unconstitutional practices ICE is deploying in our communities violate everything we believe in New York and are an assault on our democracy," Cuomo said in a prepared statement. "I demand ICE immediately cease and desist this pattern of conduct, and if they fail to do so, I will pursue all available legal recourse and commit to doing everything in my power to protect the rights and safety of all New Yorkers." 

At about 7:30 p.m., Homan issued this response:

"As a native New Yorker who began my law enforcement career as a police officer in the state, I was disappointed to learn about the Governor's grandstanding today over the issue of immigration enforcement. The Governor's comments were inaccurate and an insult to ICE's sworn law enforcement officers who conduct their lawful mission professionally and with integrity. These brave men and woman leave the safety of their homes every day to protect this great nation and our communities.
}{
What a joke! Governor Andrew Cuomo had the audacity to call out ICE agents for doing their jobs, and now he's threatening to sue them? Give me a break! The nerve of this guy, talking about "reckless and unconstitutional enforcement" practices when his own state is infested with illegal immigrants who are committing crimes left and right. Cuomo is just trying to grandstand and pander to his liberal base, but the fact is, he's failing to protect the citizens of New York. ICE agents are just doing their jobs, and they deserve respect for their hard work and dedication. Cuomo should be ashamed of himself for attacking these brave men and women. And let's not forget, this is the same guy who is protecting child rapists and allowing them to roam free in his state. What a disgrace! Cuomo needs to focus on the real problems in his state, like the rampant crime and corruption, instead of trying to score political points by attacking ICE agents. It's pathetic.
}
{non-hyperpartisan}{
News Article:

This week on "Watters' World", Jesse Watters caught up with several House Democrats who boycotted President Trump's inauguration, including Rep. John Lewis, the Atlanta Democrat who called the Republican's election "not legitimate".

Approaching Houston Congresswoman Sheila Jackson Lee, Watters asked why she was absent from the festivities:

"I have a statement on record, thank you so very much," Jackson Lee said.

WATCH: Watters Tracks Down Rep. John Lewis to Ask About Trump Criticism

'Watters' World' EXCLUSIVE: Jesse Visits Scene of Chicago Hate Crime Attack

Watters Confronts Man Who Harassed Ivanka Trump on JFK Flight

As Watters introduced himself as a Fox News reporter to Rep. John Conyers of Michigan, the Detroiter kept walking toward an open elevator, "no, sir, not right now."

"I didn't boycott it, I just didn't go to it," Conyers said.

Rep. Jerrold Nadler of New York gave a similar answer when asked, "for the reasons I stated publicly; they're on the record."

Asked if he was being divisive, the Manhattan Democrat said that if anyone wanted to unite the country, they would not have voted for Trump.

Lewis told Watters he was late and couldn't speak with him as he walked with a quick gait down a hallway.

An aide obstructed Watters' cameraman as the Congressman continued toward his destination.
}{
In a recent segment on "Watters' World", Jesse Watters spoke with several members of the House of Representatives about their absence from President Trump's inauguration. While some members provided brief statements, others declined to comment or did not respond to Watters' inquiries. Rep. John Lewis, an Atlanta Democrat who had publicly questioned the legitimacy of Trump's election, was approached by Watters but did not stop to speak with him. Lewis had previously stated that he would not attend the inauguration due to his disagreement with Trump's presidency. Other members, including Rep. John Conyers of Michigan and Rep. Jerrold Nadler of New York, also declined to comment or did not respond to Watters' questions. When asked why he did not attend the inauguration, Conyers simply stated that he "didn't go to it". Nadler, who had publicly stated that he would not attend the inauguration due to his disagreement with Trump's policies, reiterated that he did not support the President's election. Watters' segment sparked a discussion on the perceived divisiveness of the President and the responses of those who did not attend his inauguration. While some argued that the boycott was a form of protest against Trump's policies and actions, others felt that it was a divisive move that further polarized the country. Regardless of one's political beliefs, it is important to approach discussions and debates with respect and civility, and to strive for understanding and cooperation in the pursuit of common goals.
}



\genex{\ToIHeadlines}
{elections}{
News Article:

New Delhi: First reports of exit poll indicate history in West Bengal and regime change in Kerala while it’s too close to call in Tamil Nadu.

An exit poll conducted by Star News television channel has projected Congress-led United Progressive Alliance (UPA)’s victory in West Bengal and Kerala.

The survey projects Trinamool Congress - Congress alliance wiping out more than three decades old Left Front government in West Bengal.

According to the exit polls telecast in the channel on Tuesday says the TMC and Congress will win 181 and 40 respectively while Left Front will be reduced to 62 from 227.

In Kerala, the Congress-led United Democratic Front is expected to win 88 while CPM-led Left Democratic Front will win 49.

In Tamil Nadu, AIADMK is projected as winning 110 and ruling DMK will have 124 seats in the 234 member assembly.

In a close contest, Tarun Gogoi led Congress has a chance to create history in Assam by returning to power for a third consecutive term,post poll surveys predict. The Headlines Today poll predicts 44 of the 126 seats for Congress in Assam,with potential ally AUDF slated to win 14.
}{
Exit Polls predict regime change in Kerala, West Bengal; Tight contest in Tamil Nadu
}
{sports}{
News Article:

India's teen sensation P V Sindhu stormed into the third round of the World Badminton Championship with a straight game victory over Olga Golovanova of Russia in women's singles match here yesterday

Copenhagen: India's teen sensation PV Sindhu stormed into the third round of the World Badminton Championship with a straight game victory over Olga Golovanova of Russia in women's singles match here yesterday.

PV Sindhu

The men's doubles pair of Manu Attri and Sumeeth Reddy B stunned 15th seeded Japanese duo of Hirokatsu Hashimoto and Noriyasu Hirata 21-19 21-19 in 44 minutes to advance to the third round.

Sindhu, seeded 11th, took 40 minutes to prevail over her Russian opponent 21-12 21-17 in the second round match at the Ballerup Super Arena here.

She will next take on sixth seeded Yeon Ju Bae of Korea. Sindhu won a total of 42 points as compared to 29 by the Russian girl.

The world No. 12 from Hyderabad looked a bit rusty to start with and was initially trailing in the opening game.

She was playing the catching-up game till 10-10 before Sindhu managed to reel off four consecutive points and surge ahead to 14-10.

There was no looking back after that, as Sindhu swiftly sealed the game in her favour with Golovanova earning just two more points.

In the second game, the Russian got her act together and opened up a big lead, moving up to 11-6 at the break.
}{
15-year-old PV Sindhu creates history, enters World Badminton Championship 3rd round
}

\genex{\AGNews}
{World}{
News Article:

SANTIAGO Nov 14 A magnitude 6.1 earthquake hit central-north Chile on Wednesday, the U.S. Geological Survey said, causing buildings to sway in capital Santiago but causing no damage.

The USGS said the quake struck on the coast, 54 miles (88 km) north of Coquimbo, at 4:02 p.m. (1902 GMT). It was initially reported as a magnitude 6.0 quake.

Chile's state emergency office, Onemi, said there were no reports of injuries, disruptions to basic services or infrastructure damage as a result of the tremor.

The navy said the quake did not meet the conditions needed to trigger a tsunami.

World No.1 copper producer Codelco said its operations were normal, and no mines in the metal-producing country reported damages.

Seismically active Chile was battered in February 2010 by an 8.8 magnitude quake and ensuing tsunami, which caused billions in damage and killed hundreds.

}{
A magnitude 6.1 earthquake hit central-north Chile on Wednesday, causing buildings to sway in capital Santiago but causing no damage, the USGS said.
}
{Sci/Tech}{
News Article:

MONTREAL - Canadian astronaut David Saint-Jacques is bracing for some tense moments during next Monday's launch of the Soyuz rocket that will send him and two others to the International Space Station.

On Oct. 11, a rocket failure forced a Soyuz capsule carrying two astronauts to abort and make an emergency landing. Russia suspended all manned space launches pending an investigation before giving the green light Nov. 1.

Saint Jacques spoke to reporters today from the launch site in Kazakhstan where he's in quarantine. He said the most dangerous part of the six-month mission is the 10-minute Soyuz launch and the six hours that follow before docking.

He said half of the last two-and-half years of training was dedicated to his role as Soyuz co-pilot for the trip to the space station.

Saint-Jacques, 48, said once at the space station he'll be able to focus on work and life aboard the outpost.

The Canadian astronaut's first space voyage, with NASA astronaut Anne McClain and Oleg Kononenkoof of the Russian space agency Roscosmos, was originally scheduled to launch Dec. 20. It was moved ahead after Russian authorities concluded their investigation into the failed launch.
}{
commentary Saint-Jacques says the Soyuz launch is the most dangerous part of the six-month mission but he's been training for it for half of the last two-and-a-half years.
}


\genex{\Category}
{\detokenize{camera_photo}}{
Product details:

MSD ASY17296 Wiring Harness

Factory Terminated

MSD Ignition 6ALN (6430) Wiring Harness ASY17296

New harness with factory crimped connections a perfect item to have as an extra or replacement.

}{
I have a MSD Ignition 6ALN (6430) and this wiring harness is a great addition. The factory terminated connections are perfect and make installation a breeze. The harness is well made and will definitely last.

I have been using it for a few weeks now and it has performed flawlessly. Highly recommend it to anyone looking for a high quality wiring harness.
}
{apparel}{
Product details:

Conair Travel Smart 3-Dial Combination Lock (6 Pack)

Safely Locks Your Luggage With Tsa Recognized Locks Comes With A Red Inspection Status Feature Travel Sentry Approved
}{
I purchased this lock for a business trip to Europe.  I've used it several times already and so far, so good. The TSA inspectors have been able to open it without a hitch, which is great!
}


\genex{\Humor}
{humorous}{
Product details:

Keeping Up With The Kardashians - Season 5 Region

E! brings viewers back into the sensational and completely unpredictable world of one of televisions favourite families with the fourth season of the hit series Keeping Up with the Kardashians. This season, Khlo and her husband, Los Angeles Lakers star, Lamar Odom, begin their new lives as a married couple. Meanwhile, Kourtney is pregnant and back together with Scott, and Kim and her boyfriend, NFL player Reggie Bush, work on their relationship. Through all this change, the Kardashian family must find ways to evolve and keep their close-knit ties.
}{
Does anyone else notice that this is the only show where you can hear people say things like, "OMG I am totally exhausted, I need to take a nap.? And it's not considered weird?
}
{\detokenize{non_humorous}}{
Product details:

Coleman Camper's Storage Tote

Securely organize, store and carry camping gear

23" x 14" x 17"

Lightweight 600D polyester construction

Multiple pockets and compartments

Convenient carry strap

The Coleman Camper's Storage Tote is a great way to store, organize and carry your gear -- use it in the garage, in the trunk or RV, and at the campsite. Constructed of reliable, lightweight 600D polyester, the Camper's Storage measures 23" long, 14" wide and 17" tall. It features multiple pockets and compartments for organizing all your gear, and a convenient carry strap so it's easy to bring along.
}{
is this product big enough for all the camping gear for a family of four?
}


\genex{\Polarity}
{positive}{
Product details:

Birth Of Soul 1

No half measures here, but a real statement of intent, particularly with that Volume One tag. 28 cuts that helped define that thing called soul - you can argue the point, but you can't argue the music. Sublime.
}{
This album was the beginning of my love affair with soul music. These tracks are timeless and still sound amazing today.
}
{negative}{
Product details:

Beware! the Lies of Satan

It's high time we pull the covers off the enemy and expose him for who he really is, the murderer, destroyer, their, deceiver and biggest liar of all time. In this book , you will begin to discover the truth about our heavenly Father and how you can stand in victory over the devil.
}{
The book does not live up to it's promise.  There is no revelation of truth about our heavenly father, or any insight into Satan's lies.  It is simply a polemic diatribe against Satan, with no concrete solutions to any of life's problems.
}

\section{Data Preprocessing}
\label{sec:preprocessing}

\begin{table*}[!t]
\centering
\scriptsize{
\setlength{\tabcolsep}{3pt}
\begin{tabular}{
    l
    r
    L{0.71\textwidth}
}
\toprule
\bf{Country of headquarters} 
& \bf{No. articles}
& \bf{Domain} 
\\
\midrule
Antigua	& 2.6K &	\url{antiguaobserver.com} \\ 
Azerbaijan	& 70.7K &	\url{trend.az} \\ 
Bangladesh	& 28.2K &	\url{thedailystar.net} \\ 
Barbados	& 5.1K &	\url{caribbean360.com} \\ 
Brazil	& 930 &	\url{thebraziltimes.com} \\ 
China	& 10.7K &	\url{chinadigitaltimes.net}, \url{china.org.cn} \\ 
Colombia	& 22.9K &	\url{colombiareports.com}, \url{insightcrime.org} \\ 
Costa Rica	& 18.9K &	\url{ticotimes.net} \\ 
Cuba	& 1.6K &	\url{escambray.cu} \\ 
Cyprus	& 13.2K &	\url{cyprus-mail.com}, \url{dailyforex.com} \\ 
Czech Republic	& 1.2K &	\url{praguepost.com} \\ 
Egypt	& 43 &	\url{thedailynewsegypt.com} \\ 
Estonia	& 21.2K &	\url{err.ee} \\ 
Ghana	& 5.2K &	\url{ghanabusinessnews.com}, \url{modernghana.com} \\ 
Guyana	& 70.2K &	\url{stabroeknews.com} \\ 
Hong Kong	& 5.6K &	\url{asiasentinel.com}, \url{actionforex.com}, \url{hku.hk} \\ 
India	& 886.5K &	\url{mid-day.com}, \url{financialexpress.com}, \url{livemint.com}, \url{hindustantimes.com}, \url{indianexpress.com}, \url{mangalorean.com}, \url{vccircle.com}, \url{deccanchronicle.com}, \url{afaqs.com}, \url{bollywoodhungama.com}, \url{medianewsline.com}, \url{orissadiary.com}, \url{morungexpress.com}, \url{countercurrents.org}, \url{businessworld.in}, \url{governancenow.com}, \url{koimoi.com}, \url{milligazette.com}, \url{dayafterindia.com}, \url{truthdive.com}, \url{newstodaynet.com}, \url{centralchronicle.com}, \url{dalje.com}, \url{rtn.asia}, \url{realbollywood.com}, \url{mutiny.in} \\ 
Indonesia	& 2K &	\url{thejakartaglobe.com} \\ 
Iran	& 7.2K &	\url{tehrantimes.com} \\ 
Israel	& 60.4K &	\url{jewishpress.com}, \url{ynetnews.com}, \url{palestinechronicle.com}, \url{972mag.com}, \url{defense-update.com} \\ 
Jamaica	& 96.6K &	\url{jamaica-gleaner.com} \\ 
Japan	& 2.1K &	\url{japantoday.com} \\ 
Kenya	& 158.8K &	\url{capitalfm.co.ke}, \url{nation.co.ke}, \url{theeastafrican.co.ke}, \url{standardmedia.co.ke}, \url{kbc.co.ke}, \url{businessdailyafrica.com} \\ 
Kuwait	& 16.2K &	\url{arabtimesonline.com}, \url{kuwaittimes.net} \\ 
Lebanon	& 4.9K &	\url{yalibnan.com} \\ 
Macau	& 3.4K &	\url{macaudailytimes.com.mo} \\ 
Malawi	& 2.8K &	\url{maravipost.com} \\ 
Malaysia	& 30.5K &	\url{malaysiakini.com}, \url{freemalaysiatoday.com}, \url{theborneopost.com} \\ 
Misc. Africa	& 51 &	\url{african-bulletin.com} \\ 
Misc. Asia	& 30.9K &	\url{eurasiareview.com} \\ 
Namibia	& 20.2K &	\url{newera.com.na} \\ 
Nepal	& 2.2K &	\url{thehimalayantimes.com} \\ 
Nigeria	& 336.5K &	\url{thenationonlineng.net}, \url{vanguardngr.com}, \url{thisdaylive.com}, \url{codewit.com}, \url{sunnewsonline.com}, \url{businessdayonline.com}, \url{pmnewsnigeria.com} \\ 
Pakistan	& 274.1K &	\url{nation.com.pk}, \url{dawn.com}, \url{tribune.com.pk}, \url{pakobserver.net}, \url{app.com.pk}, \url{dailytimes.com.pk}, \url{thefrontierpost.com}, \url{pakistankakhudahafiz.com}, \url{thenews.com.pk}, \url{pak1stanfirst.com}, \url{pakwatan.com} \\ 
Palestine	& 655 &	\url{intifada-palestine.com}, \url{paltelegraph.com} \\ 
Peru	& 4.6K &	\url{livinginperu.com} \\ 
Philippines	& 25.1K &	\url{sunstar.com.ph}, \url{journal.com.ph}, \url{bworldonline.com}, \url{newsbytes.ph}, \url{mindanews.com}, \url{tribwekchron.com}, \url{philstar.com} \\ 
Qatar	& 8.8K &	\url{aljazeera.com}, \url{middle-east-online.com} \\ 
Romania	& 13.3K &	\url{zmescience.com} \\ 
Saint Kitts and Nevis	& 4.6K &	\url{thestkittsnevisobserver.com} \\ 
Saudi Arabia	& 42.8K &	\url{arabnews.com}, \url{saudigazette.com.sa} \\ 
Singapore	& 112.4K &	\url{straitstimes.com} \\ 
Somalia	& 197 &	\url{mareeg.com} \\ 
Somaliland	& 4.7K &	\url{somalilandpress.com} \\ 
South Africa	& 22.9K &	\url{itweb.co.za}, \url{memeburn.com}, \url{themediaonline.co.za}, \url{news24.com}, \url{iafrica.com}, \url{mybroadband.co.za} \\ 
South Korea	& 22K &	\url{koreatimes.co.kr}, \url{yonhapnews.co.kr} \\ 
Sri Lanka	& 33.8K &	\url{lankabusinessonline.com}, \url{onlanka.com}, \url{lankanewspapers.com}, \url{groundviews.org} \\ 
Tanzania	& 7.6K &	\url{thecitizen.co.tz} \\ 
Thailand	& 11.2K &	\url{pattayamail.com} \\ 
Trinidad	& 3.2K &	\url{trinidadexpress.com} \\ 
Turkey	& 2.5K &	\url{theminaretonline.com}, \url{nationalturk.com}, \url{melodika.net} \\ 
Uganda	& 6.7K &	\url{monitor.co.ug} \\ 
United Arab Emirates	& 108.8K &	\url{emirates247.com}, \url{gulfnews.com}, \url{ameinfo.com}, \url{meed.com}, \url{7days.ae} \\ 
Venezuela	& 3.9K &	\url{venezuelanalysis.com} \\ 
Zambia	& 7.4K &	\url{lusakatimes.com} \\ 
Zimbabwe	& 26.1K &	\url{newsday.co.zw}, \url{nehandaradio.com}, \url{thezimbabwemail.com} \\ 

\bottomrule
\end{tabular}
\vspace{-1ex}
}
\caption{
News domains from underrepresented countries in \RealNews.
}
\vspace{-3ex}
\label{tab:realnews-country-headquarters}
\end{table*}

\subsection{Datasets}
\begin{itemize}
    \item \AGNews: We use \url{https://huggingface.co/datasets/zapsdcn/ag}
    \item \ToIHeadlines: we use the data from \url{https://dataverse.harvard.edu/dataset.xhtml?persistentId=doi:10.7910/DVN/DPQMQH} and filter headlines in following 10 topics: \{sports, life-style, education, entertainment, business, city, environment, tech, elections, world\}. We randomly subsample to get 5.2k rows per topic in train and 1k per topic in test. 
    \item \Humor: We use \url{https://registry.opendata.aws/humor-detection/}
    \item \IMDb: We use \url{https://ai.stanford.edu/~amaas/data/sentiment/}
    \item \SST: We use \url{https://nlp.stanford.edu/sentiment/treebank.html}
\end{itemize}

Aside from \ToIHeadlines, we use the original datasets, randomly subsampling as mentioned in \autoref{tab:tasks}.

\subsection{Corpora}
\begin{itemize}
    \item \RealNews: we use the article text field and download the data from \url{https://github.com/rowanz/grover/tree/master/realnews}.
    
    \item \RealNewsRegional{} is a subset of \RealNews{} \citep{zellers2019grover}. It includes 2.7M articles from non-US and non-EU websites. We manually check \RealNews{} websites and identified 141 regional-news websites with headquarters in 56 non-US and non-EU countries: India, Pakistan, Nigeria, Philippines, etc. The complete list is mentioned in \autoref{tab:realnews-country-headquarters}.

    \item \RealNewsIndia{} is further filtered to only include Indian news websites. We use only the ``India'' domains from \autoref{tab:realnews-country-headquarters}.
    
    \item \RealNewsDominant{} is the remaining 30.1M articles from 1063 news websites headquartered in 20 countries (of which over $75\%$ are US-based). 
    
    \item \Products: We pull the data from  \url{https://nijianmo.github.io/amazon/index.html#complete-data} and concatenate title and description.

    \item \CMUMovies: Data is obtained from  \url{https://www.cs.cmu.edu/~ark/personas/}, where we use the plot summaries file.
\end{itemize}

\section{Teacher and Student hyperparameters}
\label{sec:hyperparams}

\subsection{Teacher LLM hyperparams}

For \LLaMaFull, we use the implementation from HuggingFace: \url{https://huggingface.co/TheBloke/Llama-2-13B-fp16} and run it at half-precision. 

For \ClaudeFull, we use Claude Instant v1.2: \url{https://www.anthropic.com/news/releasing-claude-instant-1-2}

We use a batch size of 1 for all generations as we have long contexts and encountered failures with higher batch sizes. We use nucleus sampling with top-p=0.9.

\subsection{Student LM hyperparams}
We use \DeBERTaLarge{} and \DistilBERT{} models from HuggingFace: \url{https://huggingface.co/microsoft/deberta-v3-large}, \url{https://huggingface.co/distilbert/distilbert-base-uncased}

We use the same hyperparameters for \DeBERTa{} and \DistilBERT{} as \citep{yu2023large}:

\begin{itemize}
    \item \DistilBERT: Learning rate of 5e-5, \detokenize{gradient_accumulation_steps} of 1, \detokenize{batch_size} 32. We use the Adam optimizer with \detokenize{weight_decay} of 1e-4 and \detokenize{epsilon} of 1e-6. We use \detokenize{max_sequence_length} of 512.  
    \item \DeBERTa: Learning rate of 2e-5, \detokenize{gradient_accumulation_steps} of 8, \detokenize{batch_size} 4. We use the Adam optimizer with \detokenize{weight_decay} of 1e-4 and \detokenize{epsilon} of 1e-6. We use \detokenize{max_sequence_length} of 512.  
\end{itemize}

We train all students for 6 epochs. Following \citep{yu2023large}, we use warmup for 6\% of the training steps.

\subsection{Oracle model hyperparams}
To train the \DeBERTaLarge{} oracle model for Label Preservation, we use a grid search over 9 combinations: 3 learning rates \{2e-5, 5e-5, 1e-4\} by 3 batch-sizes \{1, 4, 16\} (with same graident accumulation). We train on 80\% of the \gold{} training data and use the remaining 20\% as validation.

\subsection{Retriever}

We use Contriever from HuggingFace library: \url{https://huggingface.co/facebook/contriever}.

We pass a batch-size of 512 for embedding. 
\section{Computational budget}

We run all our models on AWS Elastic Cloud Compute\footnote{\url{https://aws.amazon.com/ec2/}} using 20 p3dn.24xlarge machines to call AWS cloud services, host the retrieval index and distill student models.

\subsection{Information Retrieval}

The corpora was embedded by us and the trivial was done using the Faiss library.\footnote{\url{https://faiss.ai/index.html}} We orchestrate 80 copies of Contriever using the Ray distributed framework\footnote{\url{https://docs.ray.io/en/latest/index.html}}
to embed the \RealNews{} and \Products{} corpus in $\sim$3 hours each.

\subsection{Dataset synthesis}

In order to run \LLaMaFull{} and \ClaudeFull, we invoke AWS Bedrock\footnote{\url{https://docs.aws.amazon.com/pdfs/bedrock/latest/APIReference/bedrock-api.pdf}} using the boto3 library\footnote{\url{https://boto3.amazonaws.com/v1/documentation/api/latest/index.html}}. 

Generations were done at an AWS-account level RPM of 1600 and takes roughly 4 hours for a dataset of 8k rows.

\subsection{Student distillation}

Each \DeBERTaLarge{} student model trains for 1-3 hours on a single GPU on 8k rows. Each \DistilBERT{} student model trains in 1 hour to generate the data-map for dataset catrography.

\section{Licensing}

We use datasets that have been released in prior work with various open licenses. Specifically:

\subsection{Datasets}
\begin{itemize}
    \item \AGNews: custom license, described at \url{http://groups.di.unipi.it/~gulli/AG_corpus_of_news_articles.html}
    \item \ToIHeadlines: uses Creative Commons CC0 1.0 Universal Public Domain Dedication licence as per
 \url{https://dataverse.harvard.edu/dataset.xhtml?persistentId=doi:10.7910/DVN/DPQMQH}
    \item \Hyperpartisan: taken from SemEval 2019 Task 4, this is licensed under a Creative Commons Attribution 4.0 International License as per \url{https://zenodo.org/records/1489920}
    \item \Humor: Community Data License Agreement – Sharing – Version 1.0 licence as per \url{https://registry.opendata.aws/humor-detection/}
    \item \IMDb: \citep{maas-etal-2011-learning} does not specify a licence but has made the data available for research at: \url{https://ai.stanford.edu/~amaas/data/sentiment/}
    \item \SST: \citep{socher-etal-2013-recursive} does not specify a licence but has made the data available for research at: \url{https://nlp.stanford.edu/sentiment/treebank.html}
\end{itemize}

\subsection{Corpora}
\begin{itemize}
    \item \RealNews: custom licence as per \url{https://docs.google.com/forms/d/1LMAUeUtHNPXO9koyAIlDpvyKsLSYlrBj3rYhC30a7Ak/viewform?edit_requested=true}. The code repository is Apache Licence 2.0 as per \url{https://github.com/rowanz/grover/blob/master/LICENSE}
    \item \Products: \citep{ni-etal-2019-justifying} does not specify a licence but has made the data available for research at: \url{https://nijianmo.github.io/amazon/index.html#complete-data}. 
    \item \CMUMovies: \citep{bamman-etal-2013-learning} does not specify a licence but has made the data available for research at: \url{https://www.cs.cmu.edu/~ark/personas/}.
\end{itemize}

\end{document}